\documentclass[bst/sn-nature]{sn-jnl}%
\usepackage{xr-hyper} 
\usepackage{hyperref}
\usepackage{graphicx}%
\usepackage{multirow}%
\usepackage{amsmath,amssymb,amsfonts}%
\usepackage{amsthm}%
\usepackage{mathrsfs}%
\usepackage[title]{appendix}%
\usepackage{xcolor}%
\usepackage{textcomp}%
\usepackage{manyfoot}%
\usepackage{booktabs}%
\usepackage{algorithm}%
\usepackage{algorithmicx}%
\usepackage{algpseudocode}%
\usepackage{listings}%
\usepackage[nameinlink,capitalize]{cleveref}
\usepackage{adjustbox}
\theoremstyle{thmstyleone}%
\theoremstyle{thmstyletwo}%

\theoremstyle{thmstylethree}%

\begin{document}
\title[Article Title]{Optimal trajectory-guided stochastic co-optimization for e-fuel system design and real-time operation}

\author[1,2]{\fnm{Jeongdong} \sur{Kim}}\email{jdong96@mit.edu}
\author[1]{\fnm{Minsu} \sur{Kim}}\email{mskim77@mit.edu}
\author*[3,4, 5]{\fnm{Jonggeol} \sur{Na}}\email{jgna@ewha.ac.kr}
\author*[2]{\fnm{Junghwan} \sur{Kim}}\email{kjh24@yonsei.ac.kr}

\affil[1]{\orgdiv{Department of Chemical Engineering }, \orgname{Massachusetts Institute of Technology}, \orgaddress{\street{77 Massachusetts Ave}, \city{Cambridge}, \postcode{02139}, \state{Massachusetts}, \country{United States}}}

\affil[2]{\orgdiv{Department of Chemical and Biomolecular Engineering}, \orgname{Yonsei University}, \orgaddress{\street{50 Yonsei-ro}, \city{Seoul}, \postcode{03722}, \country{Republic of Korea}}}

\affil[3]{\orgdiv{Department of Chemical Engineering and Materials Science}, \orgname{Ewha Womans University}, \orgaddress{\street{52 Ewhayeodae-gil}, \city{Seoul}, \postcode{03760}, \country{Republic of Korea}}}

\affil[4]{\orgdiv{Graduate Program in System Health Science and Engineering}, \orgname{Ewha Womans University}, \orgaddress{\street{52 Ewhayeodae-gil}, \city{Seoul}, \postcode{03760}, \country{Republic of Korea}}}

\affil[5]{\orgdiv{Institute for Multiscale Matter and Systems (IMMS)}, \orgname{Ewha Womans University}, \orgaddress{\street{52 Ewhayeodae-gil}, \city{Seoul}, \postcode{03760}, \country{Republic of Korea}}}

\abstract{E-fuels are promising long-term energy carriers with the supporting net-zero transition. However, the large combinatorial design–operation spaces under renewable uncertainty makes the use of mathematical programming impractical for co-optimizing e-fuel production systems. Here, we present MasCOR, a machine-learning-assisted co-optimization framework that learns from global operational trajectories. By encoding design and renewable trend, a single MasCOR agent generalizes dynamic operation across diverse configurations and scenarios, substantially simplifying design–operation co-optimization under uncertainty. Benchmark comparisons against state-of-the-art reinforcement learning baseline demonstrate near-optimum performance, while computational costs are substantially lower than mathematical programming, enabling rapid parallel evaluation of design within the co-optimization loop. This framework enables rapid screening of feasible design spaces together with corresponding operational policies. When applied to four potential European sites targeting e-methanol production, MasCOR showed that most locations benefited from reducing system load below 50 MW to achieve carbon-neutral methanol production, with production costs of \$1.0–1.2 kg-1. In contrast, Dunkirk (France), with limited renewable availability and high grid prices, favored system loads above 200 MW and expanded storage to exploit dynamic grid exchange and hydrogen sales to the market. These results underscores the value of the MasCOR framework for site-specific guidance, from system design to real-time operation.}

\maketitle

\refstepcounter{section}
\section*{Introduction}\label{Intro}
Decarbonization has emerged as a central pillar of global climate policy, with commitments expanding rapidly over the past decade. Current strategies largely focus on transitioning the power sector to renewable energy \cite{iea2022international}. However, renewable energy sources are inherently intermittent, have low energy density, and are unevenly distributed across regions \cite{li2024renewable,nijsse2023momentum,smith2025cost}. These limitations have underscored the need for long-term, high-energy-density energy carriers, motivating a growing interest in electrofuels (e-fuels)--synthetic fuels produced via electrified conversion \cite{millinger2021electrofuels,sherwin2021electrofuel}. This interest has already translated into action: several countries have commissioned large-scale demonstration projects exceeding 10 MW \cite{wulf2020review}. Furthermore, they can serve as carbon-neutral alternatives to fossil fuels in hard-to-abate sectors--such as chemicals, petrochemicals, and shipping--which together account for nearly 37\% of global final energy consumption \cite{kanchiralla2024role,ipcc2014climate}. Various e-fuel production pathways have been proposed, which begin with renewable-energy-powered electrolysis \cite{daiyan2020opportunities}. One promising e-fuel is e-methanol, produced via the power-to-methanol pathway, in which green hydrogen is catalytically combined with captured carbon dioxide \cite{daggash2018closing,li2024solar}. With a high carbon capture efficiency of 1.4--1.8 kg CO$_2$/kg MeOH \cite{pakdel2024techno} and compatibility with existing engines and chemical infrastructure, e-methanol is considered a promising candidate for decarbonizing hard-to-abate sectors. For instance, the maritime shipping sector, responsible for approximately 3\% of global CO emissions, is beginning to adopt e-methanol as a low-carbon alternative fuel \cite{faber2020fourth, otto2015closing}. Similarly, the chemical industry is exploring e-methanol as a sustainable feedstock to reduce reliance on fossil carbon \cite{lopez2023fossil}. 

Despite the growing acceptance of e-fuels in hard-to-abate sectors, most existing e-fuel production systems still face critical technical and operational limitations, particularly when coupled with intermittent renewable power \cite{zhang2017life}. Methanol synthesis reactors, for instance, require stable temperature, pressure, and feedstock flow, and thus operate most efficiently under steady-state conditions \cite{sternberg2015power}. Thus, for sustained practical application, the use of energy storage systems (ESS)--such as battery energy storage systems (BESS) and compressed hydrogen tanks (CHTs)--coupled with backup grid supply have to be considered to ensure seamless operation even in the face of supply variability \cite{fulham2024managing,qi2023strategies,mucci2023cost}. However, this configuration negatively impacts both the economic and environmental feasibility of these pathways. For example, installing BESS can reduce renewable curtailment by approximately 2–10\% but increase production costs by up to 60\%, raising them to \$1,670 per ton of MeOH \cite{nizami2022solar,fasihi2024global}. Conversely, incorporating steady grid power can lower production costs by up to 50\% but result in substantial carbon emissions of up to 8,000 kg CO$_2$ per ton of MeOH \cite{fulham2024managing}. These trade-offs highlight the need for co-optimization of system design (e.g., ESS size, fuel production capacity, etc.) and for operational strategies that minimize production cost with net-negative carbon emission under region-specific renewable uncertainty. 

Three key research gaps remain in the co-optimization of e-fuel production systems, limiting the applicability of existing approaches to real-world system design and operation. First, most existing studies represent renewable energy variability using empirical or probabilistic distribution \cite{ma2013scenario,bludszuweit2008statistical}. While such explicit representations are interpretable and tractable, they neglect the temporal correlation inherent in renewable profiles, limiting the fidelity and generalizability of the representation of uncertainty \cite{chen2018model}. Co-optimization is therefore typically formulated as a scenario-based bilevel problem, in which the first stage determines system design and the second stage optimizes hourly operation \cite{liu2023pathway,guo2023co}. The second gap lies in the reliance on deterministic operational solutions for second-stage optimization. While it guarantees optimality, its deterministic nature limits its applicability in real-time operation, where full-horizon information is inherently unavailable. Third, co-optimization requires solving a large number of second-stage operational problems to quantify uncertainty in system design, resulting in high computational cost and scalability issues \cite{heitsch2009scenario}. Thus, there remains a need for a new framework capable of rapidly and in parallel solving second-stage operational problems while accurately capturing the distribution of objectives and constraint violations under renewable uncertainty.
This study addresses the research gaps by introducing two machine-learning (ML) models: a generative model to capture temporal uncertainty in renewable generation and an optimal operational trajectory-driven agent model that addresses second-stage operational problems. The generative model captures region-specific temporal patterns from historical renewable datasets and generates synthetic scenarios without relying on prior distributional assumptions. These scenarios support uncertainty quantification (UQ) by providing synthetic inputs for second-stage operational evaluation. In addition, deep reinforcement learning (DRL) can be used in the second stage to approximate dynamic operational policies \cite{zhang2019deep}. However, DRL requires extensive episodic training and cannot guarantee satisfaction of operational and net-carbon-negative constraints across diverse designs and renewable generation conditions \cite{kim2025assessing}. To overcome this limitation, we developed a generalized operational agent model that directly learns from optimal solutions obtained from presolved linear programming (LP) problems. Once trained, this model can replace second-stage deterministic optimization and enable fast, parallel evaluation of dynamic operational problems under given synthetic scenario sets and first-stage design conditions.

\section*{Results}\label{result}
\subsection*{MasCOR framework for co-optimization of e-fuel systems}
\begin{figure}[!h]
\centering
\includegraphics[width=1.0\textwidth]{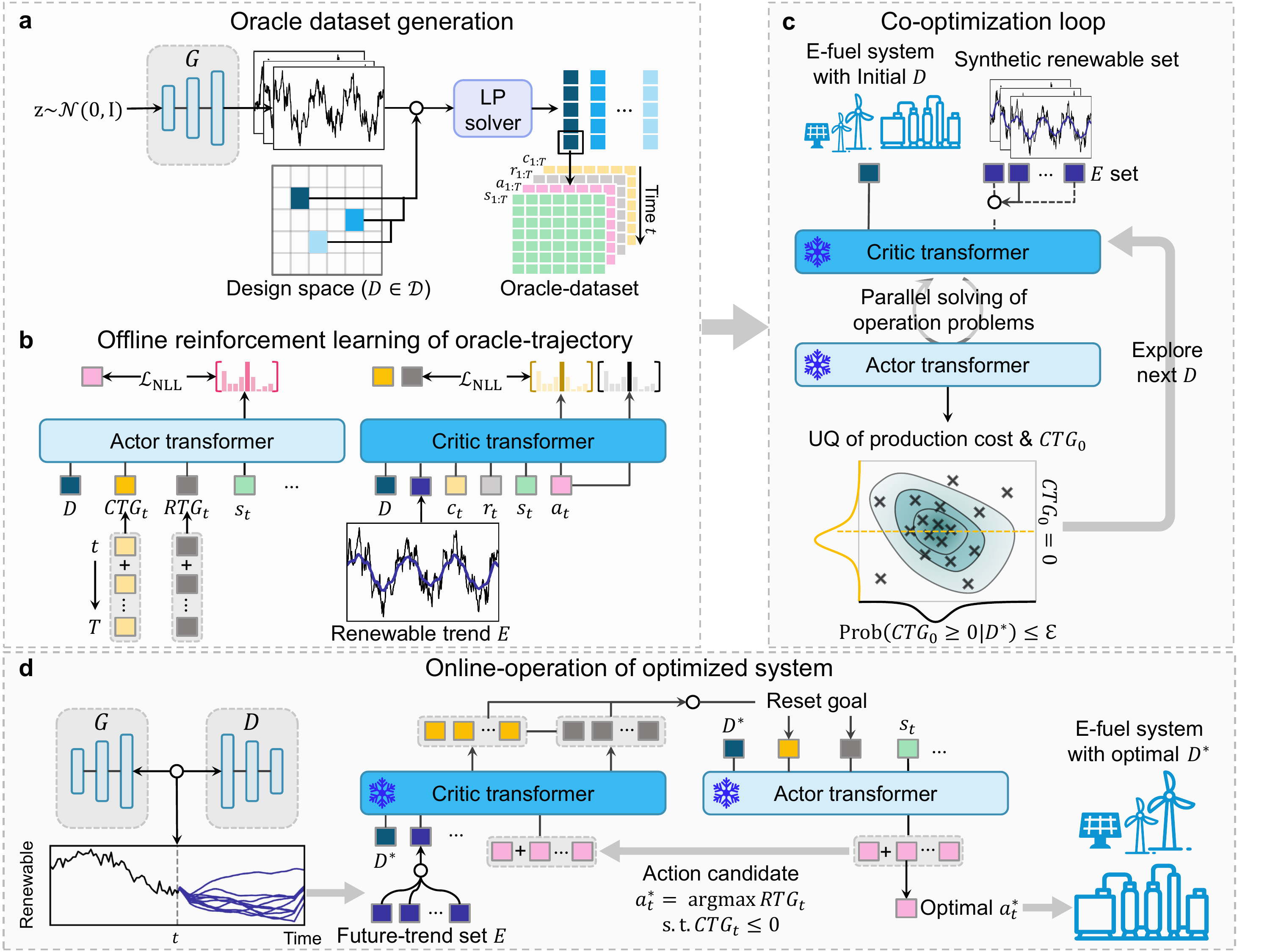}
\caption{\textbf{MasCOR: ML-assisted stochastic co-optimization framework for renewable power management systems}. \textbf{a}, Generation of an oracle dataset by combining a renewable scenario generative model with LP-based operational optimization across sampled designs. \textbf{b}, Training of transformer-based actor–critic agents to learn optimal operational policies and two future performance metrics, cumulative profit (return-to-go, $RTG_t$) and net carbon emissions (carbon-to-go, $CTG_t$), conditioned on renewable trends $E$ and system design $D$. \textbf{c}, MasCOR infers distributions of profit and net carbon emissions ($CTG_0$) for each design using the generative model and trained agent, selecting the design–policy pair that minimizes production cost and net carbon emissions while satisfying a chance constraint on positive carbon emissions ($CTG_0>0$). \textbf{d}, MasCOR generates future renewable trend sets $E$ and performs online operation conditioned on these trends and the optimal design $D^*$, ensuring robust performance under uncertainty. Details of the architecture, algorithms, and benchmark comparisons of the two ML models are provided in \textcolor{blue}{Supplementary Notes 2, 3}.}\label{fig1:mascor}
\end{figure}

We developed MasCOR (\cref{fig1:mascor}), an ML-assisted framework that co-optimizes the design (\textcolor{blue}{Supplementary Fig.~S1}) and dynamic operation of e-fuel production systems while enforcing net-negative carbon emission under region-specific temporal uncertainty in renewable generation. MasCOR represents temporal uncertainty using monthly renewable scenario sets sampled from a generative model, and it learns hourly operational policies from an oracle dataset comprising optimal operation trajectories across diverse designs and renewable trends, extracted directly from LP solutions. In this study, the hourly operational problems of e-methanol production were formulated as LP problems (\textcolor{blue}{Supplementary Note 1}) using technical parameters and region-specific carbon taxes, as listed in \textcolor{blue}{Supplementary Tables S1 and S2}. 

\begin{figure}[!h]
\centering
\includegraphics[width=1.0\textwidth]{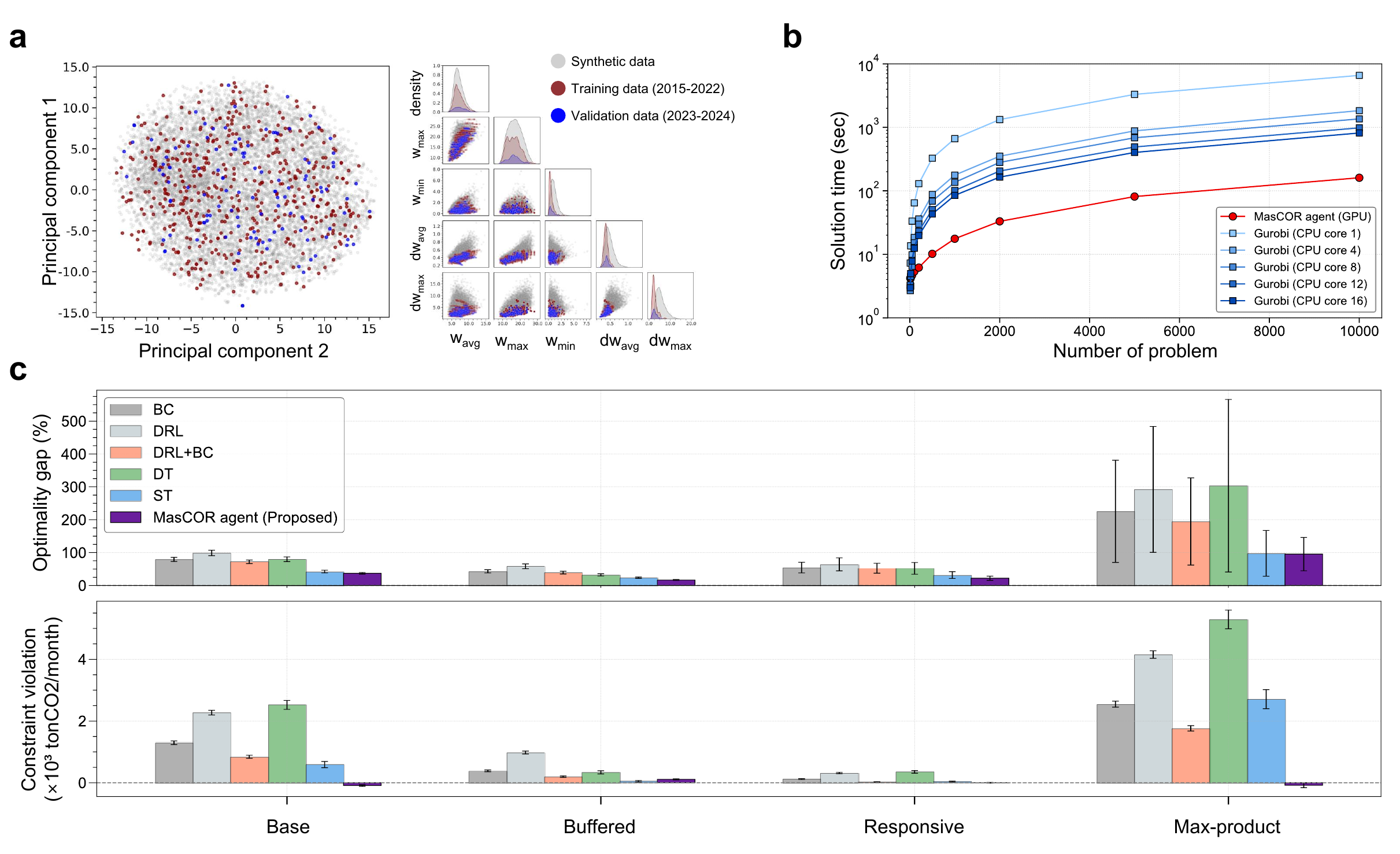}
\caption{\textbf{Performance evaluation of the MasCOR framework for renewable data generation and dynamic operational problem solving}. \textbf{a}, T-SNE embedding of synthetic renewable profiles generated by the generative model, compared with training and validation wind datasets, along with distributions of key operational features. \textbf{b}, Computational efficiency of the proposed MasCOR agent compared with the benchmark LP solver (Gurobi) for parallel LP problem solving. \textbf{c}, Dynamic operational performance comparison of benchmark data–driven models and the proposed MasCOR agent across different design categories. Details of computational efficiency and benchmark comparison experiments are provided in (\textcolor{blue}{Supplementary Notes 3.4})}\label{fig1.2:mascor}
\end{figure}

We first trained a Wasserstein generative adversarial network with gradient penalty (WGAN-GP) \cite{chen2018model, gulrajani2017improved} to generate monthly 576-hour wind speed scenarios, using the hyperparameters and network architecture listed in \textcolor{blue}{Supplementary Tables S3 and S4}. Wind speed data were collected from four target regions for training and validation; the geographical locations are listed in \textcolor{blue}{Supplementary Table S5}. An auxiliary loss enforced alignment with key statistical features of the real data, including mean power and fluctuations (\textcolor{blue}{Supplementary Note 2.1}). The resulting discriminative score remained below 0.05 across all four target sites, indicating the model’s ability to reproduce temporal dynamics with minimal discrepancy between real and synthetic data (\textcolor{blue}{Supplementary Note 2.3; Supplementary Table.~S6}). Distributional comparisons further confirmed that the generative model produces a broad range of scenarios that span both training and validation datasets while preserving key feature distributions (\cref{fig1.2:mascor}, a). Full comparisons across the different target regions are provided in \textcolor{blue}{Supplementary Figs.~S2 to S5}

After training, the WGAN-GP was used to generate monthly operational problem instances under synthetic renewable scenario sets (\cref{fig1:mascor} a). Each instance was paired with randomly sampled designs for BESS and CHT storage capacities, proton exchange membrane electrolysis cell (PEMEC) load capacity, and methanol production capacity. In addition, each instance was solved using an LP solver to maximize operational profit while satisfying net-negative emission constraints (\textcolor{blue}{Supplementary Note 1.2}). The resulting optimal solutions were then postprocessed into an oracle dataset, defined as a collection of trajectories comprising state $s_t$, action $a_t$, cost ($c_t$; carbon emissions), and reward ($r_t$; profit), and future cumulative cost and return metrics (cost-to-go, $CTG_t$; return-to-go, $RTG_t$) (\textcolor{blue}{Supplementary Note 3.1}). 

Offline reinforcement learning (RL) was performed via supervised learning on the oracle dataset using a transformer-based agent within the MasCOR framework. This agent adopts an actor–critic architecture \cite{konda1999actor}: the actor selects the next action based on $CTG_t$ and $RTG_t$ target goals, while the critic predicts future $CTG_t$ and $RTG_t$ under the selected action and updates the actor’s goals. Conditional tokens that encode the system design ($D$) and the renewable trend ($E$) enable the agent to infer policies across varying design configurations and renewable scenarios (\textcolor{blue}{Supplementary Notes 3.2, 3.3}). The proposed model offers three key advantages over conventional online RL. First, conventional online RL involves environment interaction, which is often unstable under operational constraints and inefficient when a wide design space is explored; learning directly from presolved LP solutions circumvents the need for such unstable interactions (\textcolor{blue}{Supplementary Table S7}). Second, the transformer-based architecture learns complete operational trajectories rather than isolated state–action pairs \cite{wang2022bootstrapped,torabi2018behavioral}, enabling the recovery of globally optimal operational trajectories \cite{chen2021decision}. Furthermore, by tracking $RTG_t, CTG_t$, the critic helps prevent suboptimal actions and filters out actions that fail to satisfy net-negative carbon emissions constraints (\textcolor{blue}{Supplementary Algorithm.~1 and 2}). Third, the conditional tokens $D$ and $E$ enable the use of a single generalized agent model to solve operational problems across different design configurations and renewable scenarios without fine-tuning. 

Ablation study confirmed that each component of the MasCOR agent contributes to reducing training loss (\textcolor{blue}{Supplementary Fig.~S7}). Benchmark comparisons against a deterministic LP solution and baseline RL models further demonstrated the superior performance of MasCOR (\cref{fig1.2:mascor} (b), (c)). Detailed model architectures for both the benchmark and the proposed model are provided in \textcolor{blue}{Supplementary Table S7}). The proposed model achieved an average optimality gap of 42.5\%, substantially lower than that achieved by benchmark RL models (128.2\%) across diverse design settings. Specifically, in large ESS capacity settings--referred to as the buffered case--this gap narrowed to just 16.7\%, compared with 58.7\% for the benchmarks. In addition, owing to improved goal prediction and infeasibility filtering via the critic model (\textcolor{blue}{Supplementary Table S8}), the proposed model enabled superior net-negative carbon emission performance compared with the deterministic LP solution, with the system capturing 15.4 ton/month more carbon on average. By contrast, the benchmark RL models violated constraints by an average of 1,362.1 ton/month (\textcolor{blue}{Supplementary Figs.~S8 to S11}). Overall, MasCOR delivers near-optimal dynamic operation while better satisfying net-negative emission constraints.

For co-optimization, MasCOR adopts a scenario-based bilevel optimization approach \cite{sinha2017review,dempe2002foundations} (\cref{fig1:mascor} (c)). For a given design candidate, the generative model produces a set of renewable power scenarios and the agent model solves the second-stage operational problem for each scenario in parallel. The resulting distribution of production cost and net carbon emissions across these scenarios serves as the UQ of the candidate design. The UQ procedure requires solving a large set of second-stage problems within limited computational time. The agent model of MasCOR accomplishes this through a single feed-forward inference step on a highly parallelized graphics processing unit (GPU). On a single NVIDIA H200 GPU, this yielded a 0.366–0.70 order-of-magnitude speed-up relative to the central processing unit (CPU)–based LP solver as the scenario size increased from 100 to 10,000 \cref{fig1.2:mascor} (b). Specifically, the agent model solved 1,000 scenarios in 17.6 s, compared with 84.8 s clocked by the LP solver. It is this acceleration and scalability that enables co-optimization within limited computational time. Multi-objective Bayesian optimization (MOBO) then assesses these UQ results and searches the next batch of design candidates to minimize both expected production cost and net carbon emissions in the first-stage design optimization. A chance constraint on positive carbon emissions ($CTG_0> 0$) is imposed to ensure robustness under renewable uncertainty.

Once a e-fuel system is installed under a chosen or co-optimized design, operating it requires online decision-making that can adapt to renewable uncertainty. MasCOR enables such online operation without retraining or modifying the model (\textcolor{blue}{Supplementary Note 3.2; Supplementary Algorithm.~2}). However, unlike co-optimization--during which full scenario data are available--real-time operation has only the current time step’s renewable power and grid price to rely on. The renewable-trend token $E$ must therefore be inferred dynamically. As shown in \cref{fig1:mascor} (d), at each time step $t$, the generator ($G$) and discriminator ($D$) in generative model first forecasts a set of $E$ tokens conditioned on the observed renewable data. Using these tokens, the actor proposes candidate actions, each of which the critic evaluates to estimate the distribution of $RTG_t$ and $CTG_t$ under renewable uncertainty. Finally, an optimal action that maximizes $RTG_t$ while satisfying negative $CTG_t$ constraints is selected and applied to the system. Across diverse design configurations, MasCOR achieved the lowest optimality gap (42.5\%) while limiting carbon-emission constraint violations to only 37.1 ton/month, substantially outperforming benchmark RL models (128.2\% and 269.5 ton/month) (\textcolor{blue}{Supplementary Fig.~S10}). Notably, under a large ESS capacity setting, despite relying only on hourly information--unlike LP solvers that use full-horizon scenarios--MasCOR achieved a low optimality gap of 10.28\%.

\clearpage

\clearpage

\subsection*{Power-to-methanol design and operation strategy in Europe regions}
Co-optimization was conducted for four target regions—Dunkirk (France), Skive and Fredericia (Denmark), and Weener (Germany)--within the MasCOR framework (\cref{fig1:mascor} (c)), with the objective of minimizing both the expected levelized cost of methanol production (LCOM, \$/kg) and net carbon emissions (ton/month). A chance constraint with a probability of 0.5 was enforced to ensure that the co-optimized design achieves negative carbon emissions in at least half of all renewable generation scenarios.

\begin{figure}[!h]
\centering
\includegraphics[width=1.0\textwidth]{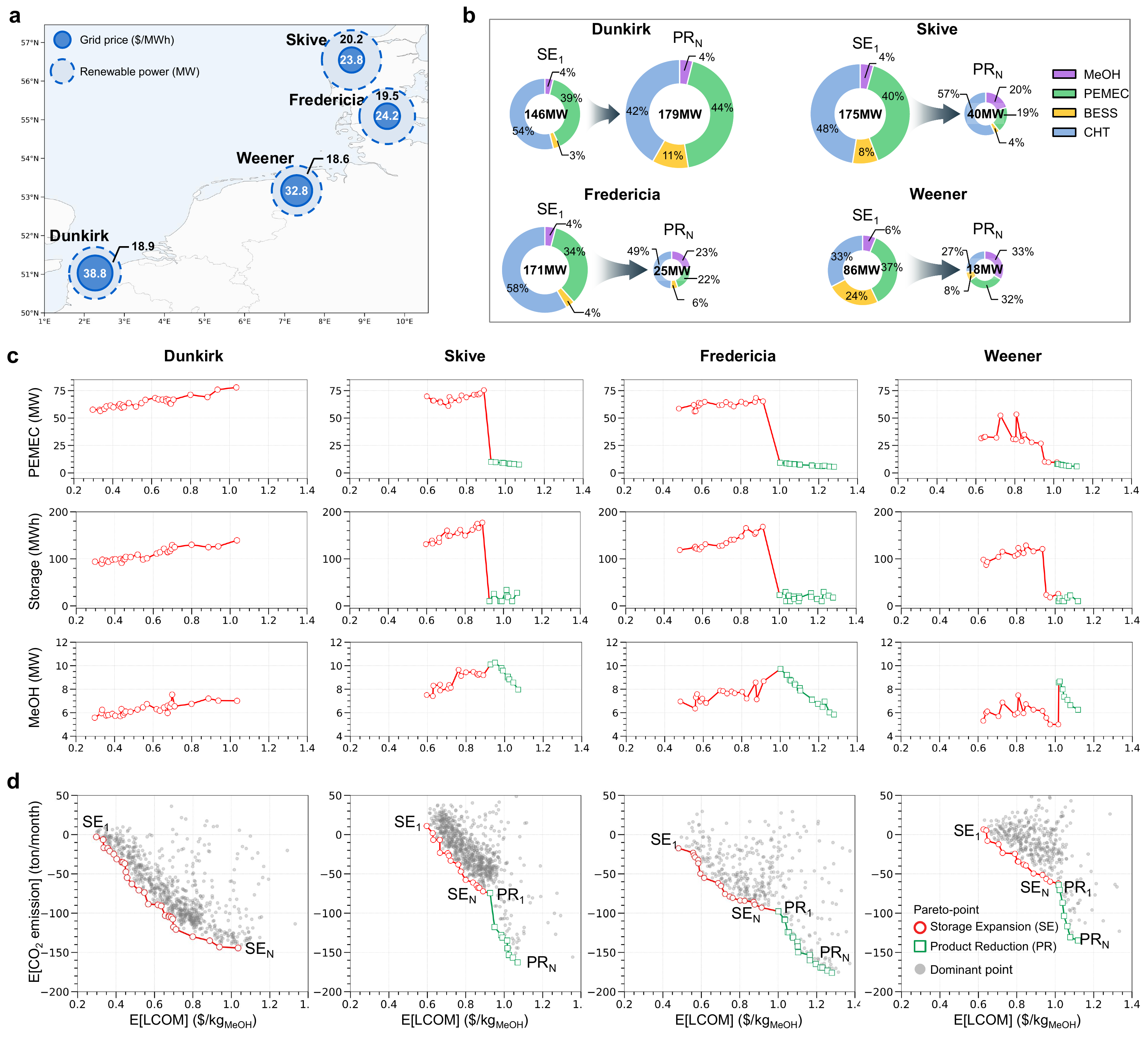}
\caption{\textbf{Regional design strategies and performance trade-offs in co-optimized power-to-methanol systems}. \textbf{a}, Differences in the average renewable power availability and grid electricity prices across Dunkirk, Skive, Fredericia, and Weener (distributional details provided in \textcolor{blue}{Supplementary Figs. S12}). \textbf{b}, Capacity shares of methanol (MeOH) production, PEMEC units, and BESS and CHT installations in representative Pareto-optimal designs, together with overall system size. \textbf{c}, Pareto-optimal capacity progression of PEMEC units, storage units (BESS and CHT), and MeOH production. \textbf{d}, Pareto frontiers of co-optimization results showing the trade-offs between the expected LCOM and net carbon emissions. MasCOR identifies two distinct design strategies for reducing carbon emission among the Pareto-optimal solutions: (i) an SE regime, characterized by increasing storage capacity, and (ii) a PR regime, characterized by decreasing MeOH production capacity under minimum PEMEC and storage capacity conditions (below 10 MW). In most regions, Pareto-optimal points shift from SE to PR as the expected net carbon emissions decrease below –50 ton/month. UQ results and full design specifications of Pareto points are provided in \textcolor{blue}{Supplementary Fig.~S13}.}\label{fig2:co-op-result}
\end{figure}

\begin{figure}[!h]
\centering
\includegraphics[width=1.0\textwidth]{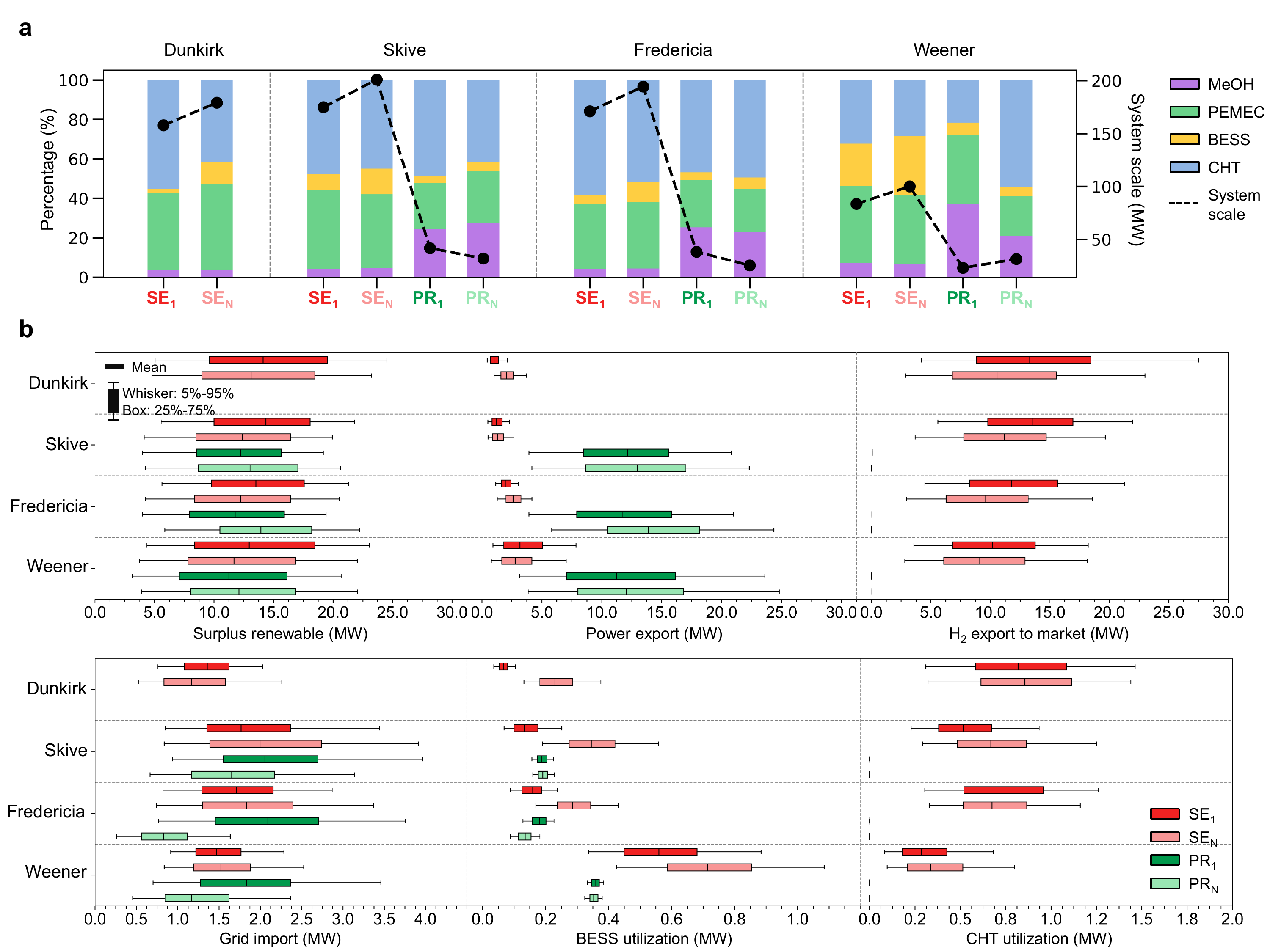}
\caption{\textbf{Regional operating strategies under different optimal design regimes}. \textbf{a}, Capacity shares (\%) of MeOH synthesis and PEMEC, BESS, and CHT units for the four representative Pareto-optimal designs illustrated in \cref{fig2:co-op-result}. The dotted line indicates system scale. \textbf{b}, Hourly operating strategies for optimal designs across regions. Distributions are obtained by averaging monthly operation profiles across 1,000 renewable scenarios using the MasCOR framework. The top panel illustrates how excess renewable power is used for either grid export or hydrogen export to the market. The bottom panel shows grid import, along with power charged to or discharged from the BESS and CHT units across the four locations.}\label{fig3:dispatch_pattern}
\end{figure}

\cref{fig2:co-op-result} shows the Pareto-optimal design patterns and expected system performance under uncertainty, revealing a clear trade-off between the LCOM and carbon emissions. As shown in \cref{fig2:co-op-result} (a), Dunkirk is characterized by relatively low renewable power availability (18.9 MW on average) and the highest grid price (38.8 \$/MWh), whereas other regions exhibit comparable renewable availability and grid price distributions. Accordingly, MasCOR identified distinct optimal design strategies across regions. Specifically, except for Dunkirk, the Pareto fronts revealed two distinct design regimes for minimizing carbon emissions: storage expansion (SE) and production reduction (PR). Dunkirk, by contrast, exhibited only the SE regime across the entire Pareto front (\cref{fig2:co-op-result} (c)). These regimes correspond to fundamentally different unit sizing, as further illustrated by their characteristic hourly operational patterns (\cref{fig3:dispatch_pattern}).

The SE regime is characterized by increased PEMEC capacity combined with large energy storage capability (BESS and CHT) (\cref{fig2:co-op-result}, (c), (d); red points), enabling surplus renewable power to be converted into hydrogen for market export. It exhibits a lower LCOM than the PR regime because market export of hydrogen provides a stable revenue stream that offsets operational costs incurred due to grid price fluctuations. For instance, as illustrated in \cref{fig3:dispatch_pattern} (b) (red box in top panel), in the case of Skive, surplus renewable power (5–25 MW) is converted into hydrogen, while only 1–2 MW is exported to the grid. However, operating an oversized PEMEC requires continuous power supply because it is the most energy-intensive unit (\textcolor{blue}{Supplementary Table S1}). During periods of renewable shortfall, the system relies heavily on the BESS and CHT to bridge the gap, as illustrated in \cref{fig3:dispatch_pattern} (b) (red box in bottom panel). Expanding this storage capacity therefore becomes pivotal for further carbon reduction in the SE regime--a pattern observed consistently across all regions (\cref{fig2:co-op-result} (c), (d)). In Skive, for example, expanding BESS and CHT capacities from 48.2 to 86.6 MWh and from 83.1 to 90.4 MWh, respectively, shifted the expected emissions from 25 to -75 ton/month and also reduced the probability of positive emissions from 0.44 to 0.35 (\textcolor{blue}{Supplementary Fig. 13 (a)}). These reductions, however, came at increased capital cost for storage, raising the LCOM and establishing a clear trade-off between cost and emissions within the SE regime.

Despite its effectiveness, the SE regime exhibits intrinsic limits in reducing carbon emissions. The oversized PEMEC increases the total system scale to beyond 150 MW (\cref{fig3:dispatch_pattern} (a)) under a renewable power capacity of only 50 MW. Therefore, when the system load remains this large, further storage expansion alone cannot eliminate the high probability of positive emissions, even at the extreme SE points in Skive, Fredericia, and Weener, where the chance probability exceeds 0.3 (\textcolor{blue}{Supplementary Fig. 13 (a)}). This limitation motivates MasCOR to transition to a distinct design regime at certain carbon emission levels.

The PR regime is characterized by minimal PEMEC and storage capacities (\cref{fig2:co-op-result} (c), (d); green points). This configuration does not permit surplus hydrogen production for market export and is strictly limited to methanol production. Thus, as shown in \cref{fig3:dispatch_pattern} (b) (green box in top and bottom panel), surplus renewable power is directly exported to the grid, with storage utilization being minimal—ranging from 0.2–0.4 MW in Skive, Fredericia, and Weener—during periods of renewable shortage. These minimal capacity settings also result in a sharp reduction in the system scale to below 50 MW as the design transitions to PR (\cref{fig3:dispatch_pattern} (a)). Because storage and PEMEC capacities are minimal in this regime, reducing methanol production capacity becomes the only viable option for further carbon emission reduction \cref{fig2:co-op-result} (c). This strategy enabled additional carbon emission reduction at the tested sites, lowering expected emissions to below -170, -170, and -150 ton/month with production capacity reductions of 10\%, 20\%, and 10\% in Skive, Fredericia, and Weener, respectively. In addition to reducing the net carbon emissions, PR effectively shifted the emission distribution and lowered the probability of positive emissions to 17\%, 8\%, and 19\% in Skive, Fredericia, and Weener, respectively (\textcolor{blue}{Supplementary Fig. 13 (a)}). However, the reduced production capacity and limited power dispatch led to higher LCOM than in the SE regime.

Overall, the discontinuous transition from SE to PR along the Pareto front reflects the intrinsic structure of the power-to-methanol system. At low LCOM, surplus hydrogen production supported by expanded storage capacity is economically optimal. However, as carbon emission constraints tighten, the high energy intensity of hydrogen production limits further emission reductions through storage expansion alone, necessitating a shift toward production downsizing. MasCOR identified this structure, producing two distinct Pareto-optimal regimes rather than a continuous design evolution in Skive, Fredericia, and Weener.

In contrast, Dunkirk exhibits only the SE regime across its Pareto front. This was driven by high grid electricity prices and strong fluctuations (20–60 \$/MWh, with hourly fluctuations of 2–6 \$/MWh), combined with limited renewable availability, as shown in \textcolor{blue}{Supplementary Fig. 12}. Under these conditions, minimizing reliance on high-priced grid power becomes essential, particularly during periods of renewable shortfall. Oversized PEMEC capacity supported by expanded storage enables the system to convert both renewable power and low-priced grid power into hydrogen, providing a stable revenue stream while ensuring continuous hydrogen supply for methanol synthesis. Consequently, storage expansion simultaneously buffers grid price volatility and supports hydrogen export, making the SE strategy consistently optimal despite its high energy intensity. For example, expanding BESS capacity from 14.7 to 64.6 MWh, with an average CHT capacity of 79.8 MWh, resulted in expected emissions of -150 ton/month with only a 16.7\% probability of positive emissions—comparable to PR-regime outcomes in the other tested regions. Moreover, Dunkirk exhibited a minimum average LCOM of 0.2 \$/kg along the Pareto front, substantially lower than the average (0.4 \$/kg) observed in the other regions. 

\clearpage

\clearpage

\subsection*{Out-of-time validation of optimal design under real-time operation}

The MasCOR framework uses a generative model to infer performance distributions of candidate system designs under region-specific renewable uncertainty (\cref{fig1:mascor} (c)). While \textcolor{blue}{Supplementary Note~2.3} confirms that the generative model accurately reproduces the temporal and statistical characteristics of both historical and future validation data, further validation is required to assess whether the inferred Pareto front and associated performance distributions hold under real-world operating conditions. To this end, we evaluated the dynamic operation of Pareto-optimal designs using a 2023–2024 validation renewable dataset excluded from both model training and co-optimization. This validation confirmed that designs identified by MasCOR exhibit robust and predictable performance under realistic renewable uncertainty.

\begin{figure}[!h]
\centering
\includegraphics[width=1.0\textwidth]{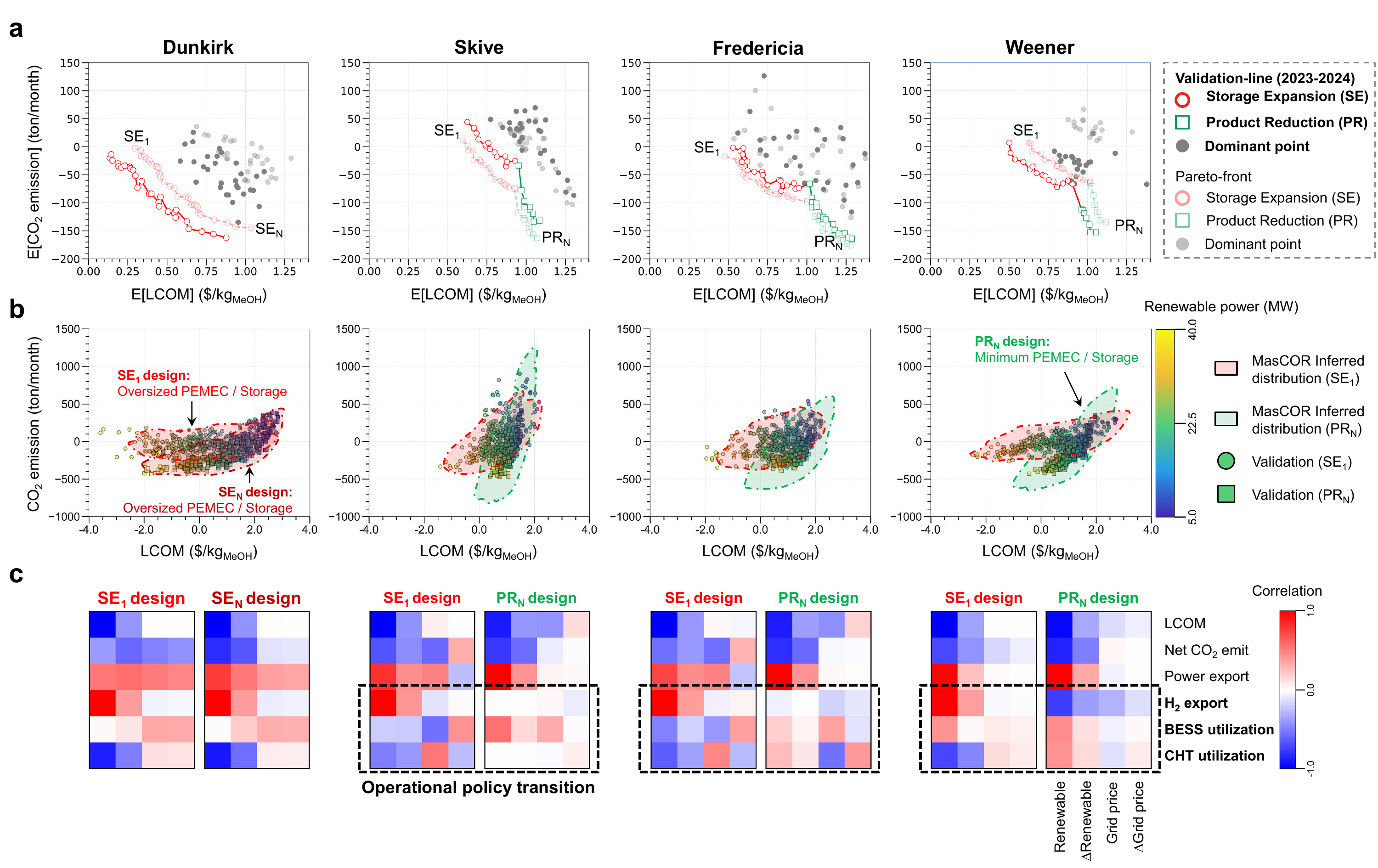}
\caption{\textbf{Validation of Pareto-optimal designs under dynamic operation}. \textbf{a}, Comparison of the MasCOR-inferred Pareto front (nonbold points) with validation results (bold points) obtained from 1,000 monthly renewable generation datasets (2023–2024). \textbf{b}, Performance distributions at the two Pareto front endpoints: dotted areas indicate 85\% confidence regions of MasCOR-inferred distributions; scatter points show validation results under varying monthly renewable power availability. Grid prices were held constant using historical data (2015–2022) to isolate renewable power uncertainty. \textbf{c}, Heat maps showing correlations between hourly renewable and grid variables and operational decisions for the two endpoint Pareto-optimal designs in each region. 
}
\label{fig4:pareto_validation}
\end{figure}

We first evaluated Pareto-optimal designs under offline operating conditions using \textcolor{blue}{Supplementary Algorithm~1}, in which renewable trend data are assumed to be known during operation. This idealized setting enables direct comparison between Pareto-optimal and dominated designs in a dynamic operational setting under a fixed renewable trend. As shown in \cref{fig4:pareto_validation} (a), the Pareto-optimal designs consistently outperformed the dominated designs and preserved the inferred Pareto front under the validation dataset. Importantly, the two design regimes identified during co-optimization, SE and PR, remained clearly distinguishable across regions.

The robustness of the inferred Pareto front is further supported by the close agreement between predicted and observed performance distributions at the two extreme Pareto designs. As shown in \cref{fig4:pareto_validation} (b), operational outcomes derived from the validation dataset closely matched the MasCOR-inferred distributions across all regions. In alignment with MasCOR’s predictions, during validation operation, an increase in renewable power availability consistently led to a reduction in both LCOM and carbon emissions, with net-negative carbon emissions achieved once average renewable power exceeded approximately 25 MW. This alignment extended to positive emissions as well: across 1,000 monthly validation operations, the observed frequency of positive carbon emissions aligned closely with the probabilities inferred by MasCOR and enforced as chance constraints during co-optimization. In Fredericia, for instance, positive emissions were recorded in 47\% and 11\% of cases for the SE1 and PRN designs, respectively, compared with inferred probabilities of 45\% and 11\%. Across all the regions, discrepancies remained below 6\%, confirming the accuracy of the inferred performance distributions under synthetic renewable scenarios. Furthermore, consistency between predicted and observed performance distributions was confirmed at the hourly operational level. The policy maps derived from the validation dataset \cref{fig4:pareto_validation} (c)) closely matched those predicted during the co-optimization loop using synthetic renewable scenarios (\cref{fig3:dispatch_pattern}). In Dunkirk, for instance, oversized PEMEC and storage capacities led to a strong positive correlation between renewable availability and hydrogen export, along with a negative correlation with grid price. These correlations between system state and operation are consistent with the inferred SE policy of converting surplus renewable electricity into hydrogen for market export (\cref{fig3:dispatch_pattern} (b), red box). Consistency between validation runs and the inferred SE policy was further demonstrated under high grid-price conditions; in line with MasCOR’s inference, increasing BESS and CHT utilization under such conditions made energy-intensive hydrogen production and continuous methanol synthesis feasible. On the other hand, the regions where both SE and PR regimes were predicted showed a clear policy transition from SE to PR during validation operations. Under PRN designs, hydrogen export became decoupled from renewable availability due to the minimum PEMEC capacity settings, while power export to the grid dominated. PEMEC operation was limited to producing only the hydrogen required for MeOH production (\cref{fig3:dispatch_pattern} (b), green box).

\begin{figure}[!h]
\centering
\includegraphics[width=1.0\textwidth]{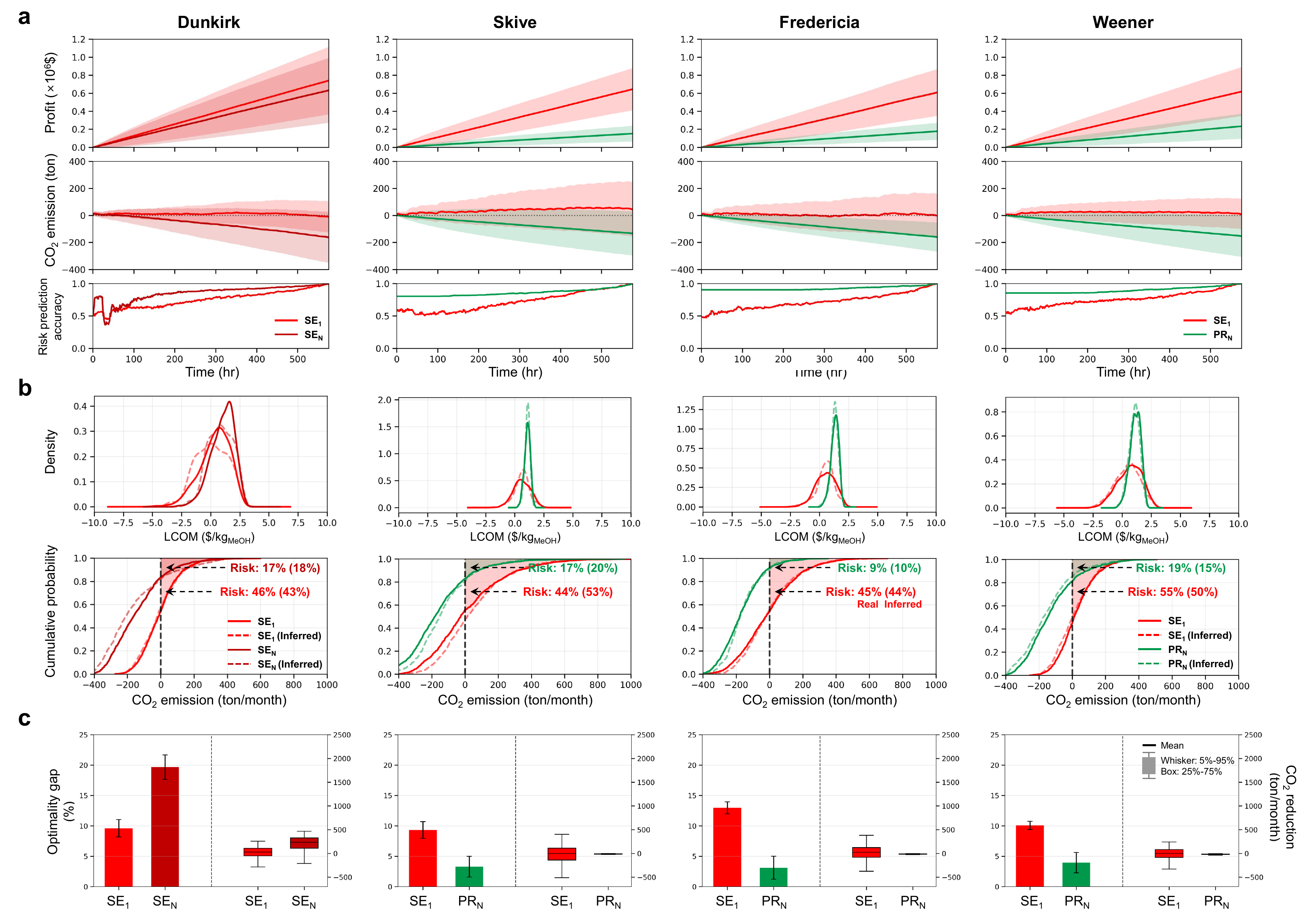}
\caption{\textbf{Online operational performance of Pareto-optimal designs across regions}. 
\textbf{a}, Trajectories of cumulative profit and net carbon emissions for the two extreme Pareto-optimal designs identified in \cref{fig2:co-op-result}. Shaded regions indicate the envelope of trajectories obtained under different validation datasets, while solid lines denote ensemble-averaged trajectories. The risk prediction accuracy charts show the ability of MasCOR’s critic component to predict the occurrence of positive net carbon emissions. \textbf{b}, Distributions of online operational outcomes compared with those inferred by MasCOR. The agreement between inferred and realized risk curves indicates the accuracy of positive carbon emission risk prediction. Close agreement indicates accurate risk estimation under renewable uncertainty. \textbf{c}, Average optimality gaps in monthly operational profit and carbon emission reduction, obtained by comparing MasCOR-based online operation with globally optimal solutions obtained from a deterministic LP solver.}
\label{fig5:pareto_validation}
\end{figure}

Following offline validation, we evaluated online operation, where the MasCOR agent was made to operate under partial observability, receiving only hourly renewable power and grid-price data (\cref{fig1:mascor}, (d)). Sets of renewable trends were dynamically forecast using the generative model and incorporated into the agent’s decision-making process using \textcolor{blue}{Supplementary Algorithm~2}. Despite limited information guiding operation, online operational outcomes remained consistent with the inferred performance distributions. As shown in \cref{fig5:pareto_validation} (b), the distributions of online operational results closely aligned with those predicted by MasCOR during co-optimization, particularly for Skive, Fredericia, and Weener. Moreover, the proportion of positive carbon emission events observed during online operation matched the inferred probability--denoted as ‘risk’ in online operation--confirming that MasCOR accurately captures both the expected performance and the emission risk associated with real-time operation.

A key advantage of the MasCOR framework is critic-based risk prediction, the prediction of the probability of positive carbon emissions. During online operation, the critic continuously predicts the expected cumulative profit ($RTG_t$) and carbon emissions ($CTG_t$). These predictions enable the MasCOR agent to prioritize actions with a higher probability of achieving profit maximization while satisfying carbon constraints. Because the co-optimization limits the probability of positive carbon emissions rather than enforce strict negativity at all times, early detection of this emission risk is critical. As shown in \cref{fig5:pareto_validation} (a), risk prediction accuracy exceeded 0.7 in early stages of online operation and converged toward unity as additional operational data accumulated. Notably, for PR$_N$ designs in Skive, Fredericia, and Weener, accuracy exceeded 0.9 within the first 24 hours of operation, enabling reliable early identification of positive emission risk even in the absence of full-horizon information.

Finally, we benchmarked online operational performance against deterministic LP solutions that assume complete foresight of renewable power and grid prices over the full planning horizon. As shown in \cref{fig5:pareto_validation} (c), MasCOR achieved near-optimal performance while often delivering superior carbon reductions. For Skive, Fredericia, and Weener, SE$_1$ designs achieved optimality gaps of below 10\% in relation to the LP benchmark while reducing emissions by approximately 500 ton/month. PR$_N$ designs achieved even smaller optimality gaps ($< 5\%$) with comparable emissions. In Dunkirk, the SE$_N$ design exhibited a larger optimality gap of approximately 20\% but achieved the largest emission reduction—approximately 1,500 ton/month—relative to the LP benchmark. These results demonstrate that MasCOR enables near-optimal real-time operation while outperforming deterministic solutions in carbon mitigation despite the absence of complete foresight.

In summary, the Pareto-optimal designs and performance distributions identified by MasCOR were validated under real-time operation across diverse regions. The framework reliably preserved Pareto optimality, accurately predicted emission risk, and delivered near-optimal performance while enabling enhanced carbon reduction. This validation demonstrates that MasCOR effectively bridges system design and real-time operational decision-making under region-specific renewable uncertainty, which is not achievable with conventional deterministic optimization approaches.

\clearpage

\clearpage

\section*{Discussion}\label{discussion}
Among various e-fuels, methanol is a particularly strong candidate for these sectors due to its high carbon utilization efficiency (1.4–1.8 kg CO$_2$ per kg methanol) \cite{pakdel2024techno} and compatibility with existing industry infrastructure \cite{faber2020fourth,otto2015closing,lopez2023fossil}. However, the limited load flexibility of the methanol synthesis unit \cite{sternberg2015power} prevents seamless production using intermittent renewable power alone. In practice, backup supply from an electricity grid is used to compensate for renewable power shortfalls, risking violation of net-negative carbon emission constraints in the process. Although previous studies have explored system sizing and operational optimization to mitigate renewable intermittency \cite{liu2023pathway,guo2023co}, most rely on deterministic solutions that fail to capture temporal uncertainty across both system design and real-time operation. Here, we address this gap by co-optimizing system design and operational policies under region-specific renewable uncertainty.

The proposed MasCOR framework integrates two ML models within a two-stage optimization process. In stage 1 (design optimization), MasCOR iteratively determines the optimal capacities of the BESS, CHT, PEMEC, and methanol production units. In stage 2 (operational optimization), a trained agent solves dynamic operational problems for candidate designs in parallel across a full set of renewable scenarios. A generative ML model produces synthetic monthly renewable profiles, enabling UQ at the design stage. The agent is trained using offline RL using optimal operational trajectories generated by a presolved linear program, allowing it to learn a near-optimal operational policy. By leveraging GPU parallelization, the framework evaluates the economic and environmental performance of each design across thousands of scenarios, ultimately identifying Pareto-optimal designs and their corresponding real-time operational policies. 

The optimal trajectory-guided agent to solving dynamic operational problems offers two key advantages over a traditional LP-based and RL approach. First, it substantially improves computational efficiency. The agent evaluates a large number of scenarios in parallel on GPUs, whereas an LP solver processes scenarios sequentially on CPUs. For example, in this study, the agent simulated approximately 1,000 scenarios in just 17.4 s, a task that took 84.8 s when using the LP solver. This speed-up enables MasCOR to efficiently explore the large combinatorial space of system designs and operational policies. Second, the agent achieves near-optimal performance, with an average optimality gap of 8\% across diverse design and scenario sets, while strictly complying with the carbon emission constraint. Importantly, unlike LP-derived solutions, agent-derived policies can be directly deployed for real-time operation. In addition, the critic component of MasCOR’s agent can predict the occurrence of positive carbon emissions early, without requiring simulation of the full operational horizon. For instance, in configurations with minimum storage and PEMEC capacities (PR$_N$) across all regions, MasCOR predicted the risk of positive carbon emissions with over 80\% accuracy within the first 24 h of operation, despite the full horizon spanning 576 h. These capabilities make the proposed approach not only computationally efficient but also readily applicable to real-world systems without further modification.

We applied MasCOR to four European regions: Dunkirk (France), Skive and Fredericia (Denmark), and Weener (Germany). Our results revealed that in regions with intermediate renewable availability and moderate grid prices (Skive, Fredericia, and Weener), the Pareto front exhibits a discontinuous transition between distinct optimal design regimes. Specifically, in the low-LCOM segment of the Pareto front, expanding storage capacities (BESS and CHT) together with an oversized PEMEC effectively reduced net carbon emissions. In this storage expansion (SE) regime (\cref{fig2:co-op-result} (c)), surplus renewable electricity is converted into hydrogen for market export, mitigating the impact of grid-price fluctuations. Concurrently, increased storage utilization reduced grid dependence and associated emissions (\cref{fig3:dispatch_pattern} (b)). However, beyond a critical point on the Pareto front, further emission reduction could not be achieved through continued storage expansion alone. MasCOR instead identified a discontinuous shift to a production reduction (PR) regime, characterized by minimal PEMEC and storage capacities. In this regime, the limited PEMEC capacity restricts surplus hydrogen production for market export, thereby reducing overall system load and carbon emissions. Excess renewable power is primarily exported to the grid or stored in the BESS unit, and further emission reduction is achieved by decreasing methanol production capacity, which becomes the dominant controllable design variable. While this SE-to-PR transition was observed in most regions, Dunkirk exhibited distinct behavior. Its substantially higher average grid price (\$ 38.8/MWh) and larger hourly fluctuations (approximately \$ 4/MWh) (\textcolor{blue}{Supplementary Fig.~12}) made sustained SE with an oversized PEMEC optimal across the entire Pareto front. In this region, minimizing reliance on high-priced grid power while monetizing surplus power through hydrogen export yielded both the lowest LCOM (approximately \$ 0.2/per kg) and net-negative carbon emissions (around -150 ton/month), without requiring a transition to the PR regime.

Out-of-time validation of Pareto-optimal designs under both offline and online operating conditions demonstrated the robustness of MasCOR-inferred designs and performance distributions. For the two extreme Pareto-optimal designs in each region, real-time operation using validation datasets showed close agreement with the inferred distributions of LCOM and net carbon emissions (\cref{fig4:pareto_validation,fig5:pareto_validation}). This indicates that the synthetic scenario–based UQ in co-optimization effectively captures the full range of possible real-world operational outcomes. Furthermore, the MasCOR critic predicts future cumulative profit and carbon emissions based on observed renewable trends during operation. Monitoring these predictions enables early detection of positive carbon emission risks, with accuracy approaching 90\% across regions, without requiring full-horizon simulation. 

While the present work provides an end-to-end decision-making framework spanning system design to real-time operation, several opportunities remain for future development. First, although renewable uncertainty is explicitly modeled using a generative approach, grid price uncertainty is currently represented through random sampling from historical data. Unlike renewable generation, grid prices exhibit highly irregular behavior—including extreme spikes and negative prices—that is difficult to capture using standard generative models. Future work could incorporate explicit stochastic price models, such as Gaussian processes, to propagate price uncertainty jointly with renewable uncertainty. Second, the current framework focuses on a single e-fuel product with fixed production capacity. Extending MasCOR to multi-fuel systems—such as methanol, ammonia, synthetic methane, and hydrogen—would enable optimization of fuel portfolios and dispatch strategies under uncertainty. Such extensions could be achieved by expanding the LP formulation while retaining the existing ML architecture. These advancements would further enhance MasCOR’s applicability for guiding investment and operational decisions toward robust, carbon-negative energy systems aligned with net-zero targets.
\clearpage

\section*{Methods}\label{methods}
\subsection*{Power-to-methanol system model}
The power-to-methanol production system comprises a PEMEC, storage units (a BESS and a CHT), and a methanol synthesis unit that utilizes captured carbon. System operation is formulated as a linear program with hourly resolution, subject to power balance constraints (\cref{eq:e_balance}) and operating limits (\cref{eq:pem_const,eq:cht_const,eq:bess_const}) \cite{kim2025assessing}. The objective \cref{eq:lp_obj} is to maximize total operating revenue while satisfying a net-negative carbon-emission constraint (\cref{eq:lp_cons}).
\begin{equation}
\max \sum_{t=0}^{T-1} 
\left[
g_t \cdot (P_t^{ex} - P_t^{g})
+ (1 - f_t^{s}) \cdot H_t^\text{sur} \cdot H_{\text{price}}
- \tau_C \cdot \varepsilon_g \cdot P_t^{g}
\right]
\label{eq:lp_obj}
\end{equation}

\begin{equation}
\text{s.t.} C^{\text{net}} = 
\sum_{t=0}^{T-1} 
\left(
\varepsilon_g \cdot P_t^{g}
- \gamma_C \cdot \dot{m}_{\mathrm{MeOH}}
\right)
\le 0
\label{eq:lp_cons}
\end{equation}
where $T$ denotes the operating horizon, $g_t$ represents the time-dependent grid power price, $H_\text{price}$ is the hydrogen price, assumed constant at 5 $\$/\mathrm{kg}$ \cite{vimmerstedt2020annual}. The grid emission factor and carbon tax are denoted by $\epsilon_g$ and $\tau_C$, respectively. $C^\text{net}$ represents cumulative net carbon emissions, $\gamma_C$ is the specific carbon consumption of the methanol synthesis unit, and $\dot{m}_{\mathrm{MeOH}}$ is the methanol production capacity. The hourly decision variables $P_t^{ex}$ and $P_t^{g}$ denote the power exported to and imported from the grid, respectively, and $f_t^s$ is the storage ratio of surplus hydrogen $H^{sur}_t$. 

For given design variables $\dot{m}_{\mathrm{MeOH}}$, $C_\text{PEM}$, $C_\text{BESS}$, and $C_\text{CHT}$, hourly operational constraints are enforced through system power balance and capacity constraints.

\begin{equation}
P_t^{r} + P_t^{g} - P_t^{ex} = P_t^{c} - P_t^{d} + P_t^{\mathrm{PEM}} + P_t^{H} + \dot{m}_{\mathrm{MeOH}} \cdot SP_{\mathrm{MeOH}}, \quad \forall t
\label{eq:e_balance}
\end{equation}

\begin{equation}
0 \le P_t^{\mathrm{PEM}} \le C_\text{PEM}
\label{eq:pem_const}
\end{equation}

\begin{equation}
0 \le H_t \le C_\text{CHT}
\label{eq:cht_const}
\end{equation}

\begin{equation}
0 \le P_t^{c,d} \le \alpha \cdot C_\text{BESS}
\label{eq:bess_const}
\end{equation}
where $P_t^{r}$ denotes the renewable power supplied to the system, $P_t^{c}$ and $P_t^{d}$ represent the BESS charging and discharging powers, respectively. $P_t^\text{PEM}$ and $P_t^{H}$ denote the loads associated with hydrogen production in the PEMEC and hydrogen storage in the CHT, respectively. The methanol synthesis unit load is the product of methanol production rate ($\dot{m}_{\mathrm{MeOH}}$) and the specific power consumption ($SP_{MeOH}$). $C_\text{PEM}$, $C_\text{BESS}$, and $C_\text{CHT}$ denote the capacities of the PEMEC, BESS, and CHT, respectively, while $\alpha$ denotes the maximum charge and discharge rate. Further details on the BESS and CHT constraints, including hourly storage level updates and technical parameters of the LP formulation, are described in \textcolor{blue}{Supplementary Note 1.2}.
\clearpage

\subsection*{Co-optimization problem formulation}
MasCOR’s primary task is stochastic co-optimization \cite{xi2014stochastic} under region-specific renewable uncertainty, while constraining the probability of positive carbon emissions (\cref{fig1:mascor} (b)). To enable reliable power-to-methanolsystem design and operation, MasCOR adopts a scenario-based bilevel optimization framework \cite{sinha2017review,dempe2002foundations}. The first stage determines the system design, while the second stage solves dynamic operational problems under a given design. 

Let $\Xi_T$ denote a set of $M$ scenarios over a time horizon of length $T$, where each scenario
\[\xi_i = \{ P_t^{(r,i)}, \, g_t^{(i)} \}_{t=0}^{T-1} \in \Xi_T, \qquad i = 1, \dots, M,\]
represents a realization of renewable power $P_t^{(r,i)}$ and grid price. The scenario set is sampled from a known continuous probability density function $p(\xi)$.

For a given design vector $d = [\dot{m}_{\text{MeOH}},\alpha_\text{PEM}, C_\text{BESS}, C_\text{CHT}] \in \mathcal{D}$ (see \textcolor{blue}{Supplementary Note~3.1 for definition}), the second-stage problem solves a monthly operational optimization for each scenario $\xi$. The objective is to maximize operational profit (\cref{eq:lp_obj}) while minimizing net carbon emissions (\cref{eq:lp_cons}), subject to hourly operational constraints (\cref{eq:e_balance,eq:cht_const,eq:bess_const}), as detailed in \textcolor{blue}{Supplementary Note~1.2}. The resulting operational decisions yield the LCOM, $l(d,\xi)$, and the net carbon emission, $e(d,\xi)$. A sample-average approximation is used to estimate the expected LCOM and carbon emission, as well as the probability of positive carbon emissions. 

The first-stage problem then optimizes the system design by minimizing the expected LCOM and carbon emissions, subject to a constraint that limits the probability of positive carbon emissions to a prescribed threshold $\alpha$.

\begin{equation}
\min_{d \in \mathcal{D}}
\Big(
\mathbb{E}_{\xi \in \Xi_T}[\, l(d,\xi) \,],
\;
\mathbb{E}_{\xi \in \Xi_T}[\, e(d,\xi) \,]
\Big),
\label{eq:upper_objective}
\end{equation}

\begin{equation}
\text{s.t.} \mathbb{P}(d) \le \alpha.
\label{eq:chance_constraint}
\end{equation}

\begin{equation}
\mathbb{P}(d)
\;:=\;
\int_{\Xi_T} \mathbb{I}\{ e(d,\xi) \ge 0 \} \, p(\xi)\, d\xi,
\label{eq:chance_constraint_def}
\end{equation}
Here, $\mathbb{I}\{\cdot\}$ denotes the indicator function, defined as $\mathbb{I}\{e\}=1$ if $e \ge 0$ and $\mathbb{I}\{e\}=0$ otherwise. 

For stochastic co-optimization, we developed two ML models within the MasCOR framework: (i) a renewable scenario generation model that captures region-specific uncertainty $p(\xi)$, and (ii) an agent-based generalized operational policy model that solves the dynamic operational problem for a given a design $d$ and a scenario $\xi$. The following subsections describe the proposed approaches, while full methodological details, model architectures, and computational experiments are provided in \textcolor{blue}{Supplementary Note}.

\subsection*{Renewable scenario generative model}
Conventional approaches to sampling renewable scenarios—from historical datasets or by fitting empirical distributions (e.g., Weibull) \cite{ma2013scenario,bludszuweit2008statistical}--cannot capture the complex temporal structure of renewable uncertainty. To address this, we employed a WGAN-GP to model region-specific temporal uncertainty, owing to its ability to map noise to complex time-series data \cite{chen2018model,goodfellow2020generative}. Synthetic wind speed scenario were generated using the WGAN-GP and converted into renewable power via turbine power curves.

However, relying solely on Wasserstein distance does not ensure reproduction of the five key operational feature distributions that are critical for power dispatch: minimum wind speed $w_{min}$, maximum wind speed $w_{max}$, and average wind speed $w_{avg}$, together with the maximum  fluctuation $dw_{max}$ and its average $dw_{avg}$ \cite{dong2022data}, that are critical for power dispatch.To overcome this limitation, we incorporated a maximum mean discrepancy loss ($L_{MMD}$) as a statistical feature for the regularization of the generator $G$ within the WGAN-GP \cite{binkowski2018demystifying}.

\begin{equation}
\min_{\theta^{(G)}} \max_{\theta^{(D)}} V(G, D) =
\underbrace{
\mathbb{E}_{\tilde{x} \sim P_g}[D(\tilde{x})] - \mathbb{E}_{x \sim P_r}[D(x)]
}_{\text{Wasserstein loss}}
+
\underbrace{
\lambda \, \mathbb{E}_{\hat{x} \sim P_{\hat{x}}} \left[ \left( \lVert \nabla_{\hat{x}} D(\hat{x}) \rVert_2 - 1 \right)^2 \right]
}_{\text{Gradient penalty loss}}
+ L_D(X, Y)
\label{eq:wgan-loss}
\end{equation}

\begin{equation}
L_{\mathrm{MMD}}(X, Y) = \mathbb{E}_{X \sim P}\!\left[ k(X, X') \right] + \mathbb{E}_{Y \sim Q}\!\left[ k(Y, Y') \right] - 2\,\mathbb{E}_{X \sim P,\, Y \sim Q}\!\left[ k(X, Y) \right]
\label{eq:mmd_loss}
\end{equation}
In this equations, $G$ and $D$ denote the generator and discriminator, each with trainable parameters $\theta$. Synthetic and real scenario distribution are denoted by $\tilde{x} \sim P_g$ and $x~P_r$, respectively. $\hat{x}$ is drawn uniformly along the line between $P_g$ and $P_r$ for the gradient penalty with coefficient $\lambda$, $X$ and $Y$ are feature vector sets of synthetic and real data with densities $P$ and $Q$, respectively, while $k$ is the kernel function. \textcolor{blue}{Supplementary Note~2} provides details of the network structure, hyperparameters, and validation studies of trained generative model \cref{eq:wgan-loss}.

\subsection*{Oracle dataset construction}
We employed an offline RL approach that directly learns the sequence of optimal operating decisions conditioned on a given design, eliminating the need for iterative LP solves. Accordingly, a large oracle dataset is constructed by solving the deterministic operational linear program (\cref{eq:lp_obj,eq:lp_cons}) over a horizon of $T=576$ hours for $M=50{,}000$ sampled scenario--design pairs $(\xi_i, d_i)$.The scenarios consist of synthetic renewable power generation data sampled from a pretrained generative model $G$ grid prices sampled from historical data, while the design vectors $d_i$ were randomly sampled from the search space defined in \textcolor{blue}{Supplementary Table S6}.  

For each scenario–design pair, the optimal LP solution was postprocessed into a Markov decision process (MDP) representation, forming the oracle dataset for offline RL. Specifically, at each time step, the state ($s_t$), action ($a_t$), cost ($c_t$), and reward ($r_t$) can be defined as follows:  
\begin{equation}
s_t = (P_t^{r}, g_t, SOC_t, H_t)
\end{equation}
\begin{equation}
a_t = (P_t^{c,d}, P_t^\text{PEM}, f_t^{u}, f_t^{s})
\end{equation}
\begin{equation}
c_t = P_t^{g} \cdot \varepsilon_g
\label{eq:ct}
\end{equation}
\begin{equation}
r_t = g_t \cdot (P_t^{ex} - P_t^{g}) + (1 - f_t^{s}) \cdot H_t^{sur} \cdot H_{price} - c_t \cdot \tau_c
\label{eq:rt}
\end{equation}
where $SOC_t$ and $H_t$ denote the state of charge of the BESS and the stored mass of the CHT, respectively, $f_t^{u}$ is the utilization ratio of stored hydrogen within the CHT, $c_t$ represents the carbon emissions from grid imports, and $r_t$ denotes the operational revenue at time $t$.  

Furthermore, in the formulated operational problem, the objective is to maximize net profit while ensuring net-negative carbon emissions by the end of the operating horizon ($T=576$), rather than prioritizing instantaneous profit or emission reductions. Thus, to train policies that reflect long-term objectives and constraints, we defined the future cumulative profit and cumulative carbon emissions.
\begin{equation}
RTG_t = \sum_{t'=t}^{T} r_{t'}, \quad CTG_t = \sum_{t'=t}^{T} c_{t'}
\label{eq:rtg-ctg}
\end{equation}

The collected trajectories $\{(s_t, a_t, c_t, r_t, CTG_t, RTG_t)\}_{t=0}^{T-1}$ form an oracle dataset compatible with an MDP. Thus, learning an operational policy from this static dataset can be formulated as an offline RL problem, enabling recovery of near-optimal dynamic operation. Details of the offline RL approach are provided in the following section.

\subsection*{An actor-critic ML model for offline RL of oracle trajectories}
We adopt a decision transformer (DT) \cite{chen2021decision} as the backbone and built an actor–critic structure \cite{barto2012neuronlike,konda1999actor}: the actor $\pi_\theta$ predict the next action $a_t$ under a given historical trajectory (\cref{eq:actor}), conditioned on two goal tokens ($RTG_t$, $CTG_t$), while the critic $Q_{\theta^{'}}$ predicts next goals ($\hat{CTG}_{t+1},\hat{RTG}_{t+1}$) under a given historical operational trajectory (\cref{eq:critic}). By updating the goal with feedback received from the critic, the actor model tracks near-optimal trajectories over extended horizons. However, optimal trajectories vary drastically across system designs and renewable scenarios; therefore, conditioning the actor policy solely on $RTG_t$ and $CTG_t$ limits generalization, which is essential for co-optimization. To address this, we introduced two conditional tokens: a design token $D$ and a renewable-trend token $E$. The design token $D$ encodes the normalized design vector $d$, while the renewable-trend token $E$ compresses the 576-hr time series into 24-hr averages. By combining these conditional tokens with the goal tokens ($CTG_t$, $RTG_t$), the agent adapts its policy to diverse designs and renewable patterns, thereby improving generalization. Using this improved setup, we trained an offline actor--critic RL agent on the oracle dataset to learn distributions over actions and corresponding future cumulative profit and carbon emission goals.

\begin{equation}
\hat{a}_t \sim \pi_{\theta}\!\left(
\mu_{\theta}, \Sigma_{\theta} \mid
\{D \cup \{CTG_i, RTG_i, s_i, a_i\}_{t-K}^{t-1} \cup \{CTG_t, RTG_t, s_t\}\right)
\label{eq:actor}
\end{equation}

\begin{equation}
\hat{RTG}_t, \hat{CTG}_t \sim Q_{\theta^{'}}\!\left(
\mu_{\theta'}, \Sigma_{\theta'} \mid
\{D, E\} \cup \{s_i, a_i, r_i, c_i\}_{t-K}^{t} \right)
\label{eq:critic}
\end{equation}
Here, $\pi_\theta$ is the actor model with learnable parameters $\theta$, while $\mu_{\theta}$ and $\Sigma_{\theta}$ denote the mean and covariance of the action distribution, respectively. $Q_{\theta^{'}}$ is the critic model with parameters $\theta^{'}$, and $\mu_{\theta'}$ and $\Sigma_{\theta'}$ are the corresponding parameters of the goal distribution. \textcolor{blue}{Supplementary Note~3} provides details on training, network architecture, hyperparameters, and computational experiment studies. 

During dynamic operation, the design token $D$ remains fixed for a given system, whereas the renewable-trend token $E$ may vary depending on the operational setting. In offline operation (\textcolor{blue}{Supplementary Algorithm 1}), $E$ can be constructed either from historical renewable data reflecting seasonal trends or from synthetic scenarios. In contrast, real-time operation in practical deployments does not assume direct access to future renewable trajectories. To accommodate this constraint, generative models ($G,D$) with \textcolor{blue}{Supplementary Algorithm 2} supports dynamic forecasting of the token $E$ using only recently observed hourly renewable data, allowing the actor--critic agent to continuously adapt its operational policy under evolving renewable uncertainty.

\subsection*{MasCOR co-optimization loop}
The MasCOR framework integrates two core ML models: a pre-trained renewable scenario generative model and an actor–critic agent. During co-optimization, these models enable the parallel solutioning of the second--stage dynamic operational problem for a given design across an $M$ set of scenarios $\Xi_T$. Replacing the conventional second-stage LP approach to ML-assisted approach makes problem computationally intractable. Thus, a derivative-free approach is used for first-stage design, where candidate designs are iteratively updated on the basis of empirical distributions of LCOM and net carbon emissions. In this study, multi-objective Bayesian optimization was adopted for the first-stage problem. The two pre-trained ML models, facukutated the direct integration of existing open-source Bayesian optimization algorithms \cite{balandat2020botorch} without modification. 

The full MasCOR co-optimization loop proceeds as follows: (i) An initial batch of $N$ design candidates $\{d_i\}_{i=1}^N$ is randomly sampled. (ii) For each $d_i$, $M$ synthetic scenarios $\{\xi_j\}_{j=1}^{M}$ are generated using the generative model for renewable power and historical sampling for grid prices. (iii) The second-stage operational problems are solved in parallel using the actor-critic model in an offline operational setting (\textcolor{blue}{Supplementary Algorithm 1}). (iv) UQ is performed, repeating stpes (ii) and (iii), after expected LCOM and carbon emissions (\cref{eq:upper_objective}), along with the probability of positive carbon emissions (\cref{eq:chance_constraint}), are computed for each design. (v) The batch of designs $\{d_i\}_{i=1}^N$ and their corresponding objectives and constraint violations are fed to the Gaussian process model, which propose the next batch of candidates. Steps (i)--(iv) are repeated until a predefined iteration count is reached.
\clearpage

\section*{Data availability}
The curated dataset supporting this study—-including meteorological data, electricity price data, and pretrained checkpoints for the generative model and the MasCOR agent—is publicly available on Kaggle at \url{https://www.kaggle.com/datasets/jeongdongkim/dataset-for-mascor-framework/data}. The original meteorological data were obtained from the NASA POWER Project database (\url{https://power.larc.nasa.gov/data-access-viewer/}), and wholesale electricity price data were obtained from the Ember European wholesale electricity price database (\url{https://ember-energy.org/data/european-wholesale-electricity-price-data/}). These raw datasets can also be manually accessed from the respective sources. The oracle dataset used for MasCOR agent training was constructed using scripts available at \url{https://github.com/PSEKJD/mascor/tree/main/test/oracle-dataset-construct}. Additional data supporting the findings of this study are available from the corresponding authors upon reasonable request.

\section*{Code availability}
The source code for MasCOR is available at \url{https://github.com/PSEKJD/mascor}. 

\section*{Acknowledgements}
This work was supported by the National Research Foundation of Korea (NRF) grant funded by the Korea government (MSIT and MOE) (RS-2025-16063688). This research was also supported by the Global C.L.E.A.N.(CCU Large-scale Emission-reduction Associative Network) through the National Research Foundation of Korea (NRF) funded by the Ministry of Science and ICT (RS-2025-02373048). This research was also supported by Korea Basic Science Institute (National Research Facilities and Equipment Center) grant funded by the Ministry of Science and ICT (RS-2024-00404602). This research was also supported by Korea Institute for Advancement of Technology(KIAT) grant funded by the Ministry of Trade, Industry \& Energy(MOTIE), Korea Government (RS-2024-00436106, Human Resource Development Program for Industrial Innovation). 

\section*{Author contributions}
J.D.K. and J.G.N. conceptualized the study and the co-optimization framework. J.D.K. developed the machine-learning-assisted co-optimization framework, conducted regional case studies, and performed validation. J.D.K., M.S.K, and J.G.N. designed the computational experiments. J.G.N. and J.H.K. conceived and supervised the project. J.D.K. wrote and edited the paper. All authors discussed and commented on the manuscript.

\section*{Competing interests}
The authors declare no competing interests.

\clearpage
\section*{Supplementary Information}
\setcounter{section}{0}
\refstepcounter{section}
\section*{Supplementary Note 1. Power-to-methanol system operation simulation}\label{sec1}
\subsection{System Configuration}\label{sec1.1}
The power-to-methanol system is selected as the target system for conducting co-optimization. The system consists of a proton exchange membrane electrolyzer (PEMEC), an electrified methanol synthesis unit, and two storage units: battery energy storage system (BESS) and compressed hydrogen tank (CHT), as shown in \cref{fig1:system-diagram} \cite{du2025cost}. For a practical application, a generalized configuration of power flows among the system units, renewable power plant, and the grid is considered. The system is directly connected to a 50 MW wind power plant, with bidirectional power exchange through the grid. In this setting, surplus renewable power can either be curtailed or exported to the grid, while grid power can be imported or exported via BESS discharging. Within the system, hydrogen from both the PEMEC and CHT is supplied to the methanol synthesis unit together with captured carbon dioxide \cite{kim2025assessing}. Surplus hydrogen can be exported to external markets, providing additional revenue to offset operating costs. The hourly operation of this configuration is formulated as a linear programming (LP) problem, as described in the following section. 

\subsection{Linear programming (LP) formulation}\label{sec1.2}
The objective of the system operation is to maximize net operational profit while satisfying net negative carbon emission. In this study, as listed in \cref{tab:technicalparameter}, the technical parameters are used to model the mass, power balance, and operating constraints in linear function. 

At time step $t$, power balance among the renewable power plant, the grid, and the system units is expressed as:
\begin{equation}
P_t^{r} + P_t^{g} - P_t^{ex} = P_t^{c} - P_t^{d} + P_t^{\mathrm{PEM}} + P_t^{H} + \dot{m}_{\mathrm{MeOH}} \cdot SP_{\mathrm{MeOH}}, \quad \forall t
\label{eq:1}
\end{equation}
where $P_t^{r}$, $P_t^{g}$, and $P_t^{ex}$ denote the renewable, grid power supplied to the system, and power exported from the system to grid, respectively. $P_t^{c}$ and $P_t^{d}$ represent the BESS charging and discharging powers, $P_t^\text{PEM}$ and $P_t^{H}$ are the loads of the PEMEC and the CHT for hydrogen production and storage, respectively. The methanol synthesis unit load is given by the product of methanol production rate $\dot{m}_{\mathrm{MeOH}}$ and the specific power consumption $SP_{MeOH}$.

To produce $\dot{m}_{\mathrm{MeOH}}$, the hydrogen supplied from both the PEMEC and CHT must satisfy the hydrogen demand of the methanol synthesis unit. Surplus hydrogen $H_t^\text{sur}$ can be stored in CHT with the ratio of $f_t^s$ or exporting to the market, and its storage level $H_t$ is updated with the capacity constraints.

\begin{equation}
H_t^{\mathrm{sur}} = f_t^{u} \cdot H_t + \frac{P_t^{\mathrm{PEM}}}{SP_{H_2}} - \gamma_{H} \cdot \dot{m}_{\mathrm{MeOH}}
\label{eq:2}
\end{equation}

\begin{equation}
H_{t+1} = (1 - f_t^{u}) \cdot H_t + f_t^{s} \cdot H_t^{\mathrm{sur}}
\label{eq:3}
\end{equation}

\begin{equation}
P_t^{H} = f_t^{s} \cdot H_t^{\mathrm{sur}} \cdot SPC_{H_2}
\label{eq:4}
\end{equation}

\begin{equation}
0 \le P_t^{\mathrm{PEM}} \le C_\text{PEM}
\label{eq:5}
\end{equation}

\begin{equation}
0 \le H_t \le C_\text{CHT}
\label{eq:6}
\end{equation}
where $f_t^u$ is the utilization ratio of stored hydrogen for methanol synthesis, $SP_{H2}$ is specific power consumption for hydrogen production of PEMEC, $\gamma_H$ is specific hydrogen consumption for methanol production, $f_t^s$ is the storage ratio of surplus hydrogen, and $SPC_{H2}$ is specific power consumption for hydrogen storage, $C_\text{PEM}$ is the capacity of PEMEC, and $C_\text{CHT}$ is the capacity of storage tank. 

In case of BESS, linear constraint of BESS operation is considered to prevent battery overload. The state of charge $SOC_t$ is updated with charging and discharging efficiency and self-discharging rate, while practical limits of $SOC_t$ are applied to prevent degradation and ensure system reliability. 
\begin{equation}
0 \le P_t^{c,d} \le \alpha \cdot C_\text{BESS}
\label{eq:7}
\end{equation}

\begin{equation}
SOC_{t+1} = \left( SOC_t + \frac{\eta_c \cdot P_t^c}{C_\text{BESS}}
- \frac{P_t^d}{\eta_d \cdot C_\text{BESS}} \right) (1 - \eta_l)
\label{eq:8}
\end{equation}

\begin{equation}
0.1 \le SOC_{t} \le 0.9
\label{eq:9}
\end{equation}
where $\alpha$ denotes the maximum charge and discharge rate, $\eta_c$ and $\eta_d$ represent the charging and discharging efficiencies, $\eta_l$ is the self-discharge rate, and $C_\text{BESS}$ is battery capacity. 

Built upon the above hourly operational constraints, the objective function is defined as the monthly operating revenue, while ensuring net-negative carbon emissions for carbon-neutral methanol production. 

\begin{equation}
\max \sum_{t=0}^{T-1} 
\left[
g_t \cdot (P_t^{ex} - P_t^{g})
+ (1 - f_t^{s}) \cdot H_t^\text{sur} \cdot H_{\text{price}}
- \tau_C \cdot \varepsilon_g \cdot P_t^{g}
\right]
\label{eq:10}
\end{equation}

\begin{equation}
\text{s.t.} C^{\text{net}} = 
\sum_{t=0}^{T-1} 
\left(
\varepsilon_g \cdot P_t^{g}
- \gamma_C \cdot \dot{m}_{\mathrm{MeOH}}
\right)
\le 0
\label{eq:11}
\end{equation}
where $T$ is operating horizon, $g_t$ is grid price, and and $H_\text{price}$ is hydrogen price, which is to be constant at 5 $\$/\mathrm{kg}$ \cite{vimmerstedt2020annual}, $\epsilon_g$ is emission factor, which is fixed at $\mathrm{ton CO_2}/\mathrm{MWh}$, consistent with worldwide carbon intensity for grid electricity in 2015 \cite{iea_electricity_2025_emissions}, and $\tau_C$ is carbon tax, listed in \cref{tab:carbon-tax}, $C^\text{net}$ is overall carbon emission, and $\gamma_C$ is specific carbon consumption of methanol synthesis unit. 

\refstepcounter{section}
\section*{Supplementary Note 2. Renewable scenario generative model}\label{sec2}
\subsection{Training procedures and model architecture}\label{sec2.1}
The generative adversarial network is trained with the Wasserstein distance and gradient penalty (WGAN-GP) \cite{chen2018model, gulrajani2017improved}, while an additional maximum mean discrepancy (MMD) $L_D$ loss on key operational features is applied for training a generator to ensure realistic scenario distributions and operational characteristics \cite{binkowski2018demystifying}.

\begin{equation}
\min_{\theta^{(G)}} \max_{\theta^{(D)}} V(G, D) =
\underbrace{
\mathbb{E}_{\tilde{x} \sim P_g}[D(\tilde{x})] - \mathbb{E}_{x \sim P_r}[D(x)]
}_{\text{Wasserstein loss}}
+
\underbrace{
\lambda \, \mathbb{E}_{\hat{x} \sim P_{\hat{x}}} \left[ \left( \lVert \nabla_{\hat{x}} D(\hat{x}) \rVert_2 - 1 \right)^2 \right]
}_{\text{Gradient penalty loss}}
+ L_D(X, Y)
\label{eq:wgan-loss}
\end{equation}
where $G$ and $D$ are generator and discriminator with trainable parameters $\theta$, $\tilde{x} \sim P_g$ and $x~P_r$ are synthetic and real scenario distribution, $\hat{x}$ is drawn uniformly along the line between $P_g$ and $P_r$ for the gradient penalty with coefficient $\lambda$, $X$ and $Y$ are feature vector set of synthetic and real scenarios.

In this study, $L_D$ accounts for five key operational features—minimum wind speed $w_{min}$, maximum wind speed $w_{max}$, and average wind speed $w_{avg}$, together with the maximum  fluctuation $dw_{max}$ and its average $dw_{avg}$ \cite{dong2022data}. 

\begin{equation}
w_{\min} = \min_{0 \le t \le T-1} w_t^{r}, \quad
w_{\max} = \max_{0 \le t \le T-1} w_t^{r}, \quad
w_{\mathrm{avg}} = \frac{1}{T} \sum_{t=0}^{T-1} w_t^{r}
\label{eq:renew-feat1}
\end{equation}

\begin{equation}
dw_{\max} = \max_{1 \le t \le T-1} \left| w_t^{r} - w_{t-1}^{r} \right|, \quad
dw_{\mathrm{avg}} = \frac{1}{T-1} \sum_{t=1}^{T-1} \left| w_t^{r} - w_{t-1}^{r} \right|
\label{eq:renew-feat2}
\end{equation}

Capturing the distributions of these features is essential to the robustness of co-optimization, as they directly govern power-to-methanol system performance. The MMD is then calculated on the feature vectors using a multi-scale Gaussian kernel:
\begin{equation}
X = \{ x_1, \ldots, x_n \}, \quad
Y = \{ y_1, \ldots, y_m \}
\end{equation}

\begin{equation}
k(x, y) = \sum_{\sigma \in \Omega} \exp\left( -\frac{ \lVert x - y \rVert^2 }{ 2\sigma^2 } \right)
\end{equation}

\begin{equation}
L_D(X, Y) =
\frac{1}{n^2} \sum_{i=1}^{n} \sum_{j=1}^{n} k(x_i, x_j)
+ \frac{1}{m^2} \sum_{i=1}^{m} \sum_{j=1}^{m} k(y_i, y_j)
- \frac{2}{nm} \sum_{i=1}^{n} \sum_{j=1}^{m} k(x_i, y_j)
\end{equation}
Here, $x$ and $y$ are feature vectors, and $k$ is a multi-scale Gaussian kernel function with bandwidth $\sigma \in \Omega$. Empirically, $L_D$ is defined as the discrepancy between the similarities within synthetic features, within real features, and across synthetic and real features. 

Following adversarial loss of \cref{eq:wgan-loss}, the model is trained in an alternating manner: three updates’ steps of $\theta^{(D)}$ are performed for every update step of the $\theta^{(G)}$, where $\theta^{(G)}$ is additionally optimized with $L_D$. Parameter updates are performed using the root mean square propagation, and hyperparameters are listed in \cref{tab:gan-hyperparameter}. 

To capture temporal dynamics of renewable power, two-dimensional convolutional layers are employed. For $G$, the input noise is sampled from a standard normal distribution and converted into a synthetic scenario of length 576 hours. For $D$, a 576-length scenario is reshaped into a 24 by 24 single-channel image and passed through the network to compute a Wasserstein distance value. Detailed model structures are provided in \cref{tab:gan-network}. 

\subsection{Dataset collection}\label{sec2.2}
For training the WGAN-GP and validating the co-optimization results, a renewable dataset was collected from 1 January 2015 to 31 December 2024 at 1-h resolution. Four sites with ongoing power-to-fuel projects exceeding 10 MW were selected: Dunkirk (France), Skive and Fredericia (Denmark) and Weener (Germany) \cite{wulf2020review}. In this study, the wind-speed at 50 m as a renewable source was retrieved using the NASA POWER API \cite{sparks2018nasapower} based on the site-specific latitude and longitude listed in \cref{tab:region-lat-long}. The wind-speed data retrieved from 2015 to 2022 were used for model training, while data from 2023 to 2024 were used for validation.

\subsection{Model validation}\label{sec2.3}
For quantitative measure of similarity of synthetic wind speed data, we measured discriminative score using time-series classification model—a standard approach for evaluating the performance of data-driven generative models \cite{yoon2019time}. After training WGAN-GP, synthetic wind-speed data from generator was labeled as fake, while training dataset was labeled as real. Then, a two-layer LSTM network as classification was trained for supervised classification task via binary cross entropy loss minimization. After training with 0.0001 learning rate and 100 epochs, the discriminative score—defined as the absolute deviation of the classifier’s accuracy from 0.5—was computed on the validation dataset of real and synthetic wind-speed data. A lower discriminative score (i.e., classification accuracy close to 0.5) indicates higher similarity between real and synthetic datasets, reflecting the generator’s capability to reproduce realistic temporal patterns of wind-speed. 

\cref{tab:discriminative-score} lists the discriminative scores computed for the four target regions. All regions exhibited scores below 0.05, substantially lower than reported benchmarks dataset and other generative models, which typically show scores exceeding 0.10 \cite{yoon2019time}. These results indicate that the trained generator for all target regions successfully reproduces the temporal patterns of real wind speed data. Additionally, t-SNE analysis \cite{maaten2008visualizing} and distributional comparison of five key operational features—$p_{min},w_{max},w_{avg}, dw_{max}, dw_{avg}$ (\cref{eq:renew-feat1,eq:renew-feat2})—were performance for the training, validation, and synthetic datasets of each regions. \cref{fig2:tsne-dunkirk,fig3:tsne-skive,fig4:tsne-fredericia,fig5:tsne-weener} visualizes the similarity between the real and synthetic data distributions in the 2-D embedding space, where the broader coverage of the synthetic data indicates the generator’s capability to produce a diverse set of wind-speed scenarios.

\refstepcounter{section}
\section*{Supplementary Note 3. Agent-based generalized operational policy model}\label{sec3}
\subsection{Oracle dataset construction}\label{sec3.1}
The oracle dataset, defined as the monthly optimal operational trajectories, was constructed from a linear programming (LP) problem instance (\cref{sec1.2}). This dataset was used for offline reinforcement learning of actor-critic agent model, enabling the inference of a generalized operational policy for the power-to-methanol system under given design vectors and renewable scenario trends. 

The problem instance set was generated by sampling $M$ set of renewable power and grid price scenarios, together with system design vectors, collectively denoted as
\begin{equation}
\{\xi_i,\, d_i\}_{i=1}^M, 
\quad \text{where } 
\xi_i = \{P_t^{(r,i)},\, g_t^{(i)}\}_{t=0}^{T-1}.
\label{eq:oracle_instance}
\end{equation}
Here $\{P_t^{(r,i)},\, g_t^{i}\}_{t=1}^{T}$ represents $i$-th scenario of renewable power and grid price over the horizon $T$, and $d_i$ denotes the power-to-methanol design vector, defined in \cref{eq:design_vector}. 

In this study, the operational horizon was fixed at $T=575$~hours for monthly operation. The synthetic wind-speed scenarios $\{w_t^{i}\}_{t=0}^{T-1}$ were first generated using the trained generator $G$ (\cref{sec2.1}), and subsequently converted into renewable power $\{P_t^{r,i}\}_{t=0}^{T-1}$ through the turbine capacity-factor function, assuming an installed capacity of 50~MW. Separately, the grid price scenarios $\{g_t^{i}\}_{t=0}^{T}$ were drawn from historical datasets of the four target regions covering the period from 2015 to 2022 \cite{EMBER}:

\begin{equation}
\{w_t^{(r,i)} \}_{t=0}^{T-1} = G_\theta(z_i), 
\qquad 
z_i \sim \mathcal{N}(0, \Sigma)
\end{equation}

\begin{equation}
P_t^{(r,i)} =
\begin{cases}
0, & w_t^{(i)} \le w_{\mathrm{cut\text{-}in}}, \\
P_{\mathrm{install}}
\dfrac{(w_t^{i})^{3} - w_{\mathrm{cut\text{-}in}}^{3}}
      {w_{\mathrm{rate}}^{3} - w_{\mathrm{cut\text{-}in}}^{3}}, &
w_{\mathrm{cut\text{-}in}} < w_t^{i} \le w_{\mathrm{rate}}, \\
P_{\mathrm{install}}, & 
w_{\mathrm{rate}} < w_t^{(r,i)} \le w_{\mathrm{cut\text{-}off}}, \\
0, & w_t^{i} > w_{\mathrm{cut\text{-}off}},
\end{cases}
\label{eq:wind-power-function}
\end{equation}
where $z_i$ denotes the Gaussian noise vector, the cut-in $w_{\mathrm{cut\text{-}in}}$, rated $w_{\mathrm{rated}}$, and cut-off $w_{\mathrm{cut\text{-}off}}$ wind speeds are set to 1.5, 12, and 25 m/s, respectively \cite{kim2023revealing}, and the installed wind capacity $P_{\mathrm{install}}$ is assumed to be 50MW.

The design vectors $d_i$ defining the capacity constraints in \cref{sec1.2} were generated using Latin hypercube sampling \cite{loh1996latin} within the search space specified in \cref{tab:design_space}. Specifically, the PEMEC capacity ($C_\text{PEM}$) was determined based on the ratio $\alpha_\text{PEM}$ between the minimum load ($P_{\text{PEM}}^{\min}$) required for methanol production rate $\dot{m}{\text{MeOH}}$ and the maximum load ($P_{\text{PEM}}^{max})$) corresponding to concurrent methanol production and full CHT charging, thereby ensuring operational feasibility across all sampled design vectors: 
\begin{equation}
d_i = [\dot{m}_{\text{MeOH}}^i,\alpha_\text{PEM}^i, C_\text{BESS}^i, C_\text{CHT}^i]
\label{eq:design_vector}
\end{equation}

\begin{equation}
\begin{aligned}
P_{\text{PEM}}^{\min} &= \gamma_H \cdot \dot{m}_{\text{MeOH}}^i \cdot SP_{H_2}, P_{\text{PEM}}^{\max} = P_{\text{PEM}}^{\min} + C_\text{CHT}^i \cdot SP_{H_2} \\
C_\text{PEM}^i &= \alpha_\text{PEM}^i (P_{\text{PEM}}^{\max} - P_{\text{PEM}}^{\min}) + P_{\text{PEM}}^{\min},
\end{aligned}
\label{eq:PEM-cap-selection}
\end{equation}

Given $M = 50{,}000$ sampled problem instance $\{\xi_i,\, d_i\}_{i=1}^M$, the LP problem set was iteratively solved using the Gurobi optimizer with default setting \cite{gurobi}. Conditioned on $(\xi_i, d_i)$, the trajectory of state $s_t$, action $a_t$, cost $c_t$, and reward $r_t$ were defined as follows:
\begin{equation}
s_t = (P_t^{r}, g_t, SOC_t, H_t)
\end{equation}

\begin{equation}
a_t = (P_t^{c,d}, P_t^\text{PEM}, f_t^{u}, f_t^{s})
\end{equation}

\begin{equation}
c_t = P_t^{g} \cdot \varepsilon_g
\label{eq:ct}
\end{equation}

\begin{equation}
r_t = g_t \cdot (P_t^{ex} - P_t^{g}) + (1 - f_t^{s}) \cdot H_t^{sur} \cdot H_{price} - c_t \cdot \tau_c
\label{eq:rt}
\end{equation}
where the cost $c_t$ represents the carbon emission from grid power usage, and the reward $r_t$ denotes the hourly operational revenue at time $t$.

The carbon-to-go $CTG_t$, reward-to-go $RTG_t$ were then defined as cumulative sums of $c_t$ and $r_t$, respectively: 
\begin{equation}
RTG_t = \sum_{t'=t}^{T-1} r_{t'}, \quad CTG_t = \sum_{t'=t}^{T-1} c_{t'}
\label{eq:rtg-ctg}
\end{equation}

Finally, the oracle dataset $\mathcal{D}_{\text{oracle}}$, used for offline reinforcement learning of agent model in \cref{sec3.2}, is defined as the collection of optimal trajectories of tuples $\tau_i$ for all $i = 1, \dots, M$:
\begin{equation}
\tau_i = \bigl\{(CTG_t,\, RTG_t,\, s_t,\, a_t,\, c_t,\, r_t) \bigr\}_{t=0}^{T-1}.
\label{eq:oracle-trajectorty}
\end{equation}
\begin{equation}
\mathcal{D}_{\text{oracle}} = \{(\tau_i|\xi_i,d_i)\}_{i=1}^{M}.
\label{eq:oracle-dataset}
\end{equation}

\subsection{Offline reinforcement learning of oracle trajectory}\label{sec3.2}
We developed a generalized operational policy model conditioned on design vectors $d_i$, adopting a Decision Transformer (DT) \cite{chen2021decision} as the backbone and employing an actor–critic architecture \cite{konda1999actor}. 
The trained generative model (\cref{sec2.1}) enabled the construction of the oracle dataset $\mathcal{D}_{\text{oracle}}$, which contains optimal operational trajectories $\tau_i$ across diverse design settings and renewable scenarios specified for the target regions. 

Oracle dataset from LP solution allowed offline reinforcement learning (RL) of the conditional policy directly from expert trajectories, removing the need for environment interaction in online RL, which is often unstable under long-term constraints or varying design conditions.
Moreover, the transformer-based architecture enables learning optimal operational sequence itself and provides more reliable long-horizon policy predictions, compared with conventional single-step models such as behavior cloning (BC) \cite{wang2022bootstrapped,torabi2018behavioral}.
The performance comparison with the online RL and BC-based policies is presented in \cref{sec3.4.3} to support this statement.

The actor transformer network $\pi_{\theta}$ predicts the next action $\hat{a}_t$ given a $K$-length historical trajectory $(s_t, a_t)$.
To condition the policy on target future performance while maintaining negative carbon emissions, additional goal tokens—$RTG_t$ and $CTG_t$—defined in \cref{eq:rtg-ctg} are concatenated to the trajectory sequence.
To further enable policy generalization across different design configurations, a design token $D$, normalized design vector $d_i \in [0,1]$ within the predefined design space (\cref{tab:design_space}), is embedded as conditional token:
\begin{equation}
\hat{a}_t \sim \pi_{\theta}\!\left(
\mu_{\theta}, \Sigma_{\theta} \mid
\{D \cup \{CTG_i, RTG_i, s_i, a_i\}_{t-K}^{t-1} \cup \{CTG_t, RTG_t, s_t\}\right)
\label{eq:actor}
\end{equation}
where $\mu_{\theta}$ and $\Sigma_{\theta}$ denote the learnable mean and covariance parameters of the action distribution.   

The critic transformer network $Q_{\theta'}$ predicts the goal tokens—$RTG_t$ and $CTG_t$—based on the given design and renewable scenarios.
Unlike the actor, the critic additionally receives the renewable-trend token $E$ as conditional token, which compresses 576-hour renewable patterns $\{P_t^{(r,i)}\}_{t=1}^{T}$ into 24-hour averages, allowing the network to account for variability in renewable availability:

\begin{equation}
\hat{RTG}_t, \hat{CTG}_t \sim Q_{\theta^{'}}\!\left(
\mu_{\theta'}, \Sigma_{\theta'} \mid
\{D, E\} \cup \{s_i, a_i, r_i, c_i\}_{t-K}^{t} \right)
\label{eq:critic}
\end{equation}

\begin{equation}
E = [E_h]_{h=0}^{N}, \qquad 
E_h = \frac{1}{24} \sum_{k=0}^{23} P^{r}_{k+24h}
\label{eq:renewable-trend}
\end{equation}
where $\mu_{\theta'}$ and $\Sigma_{\theta'}$ denote the learnable mean and covariance parameters of the goal distribution, $P_k^{r}$ represents hourly renewable power. 

\subsection{Training and operation}\label{sec3.3}
\noindent\textbf{Model training} Both the actor and critic networks are trained to model the probabilistic distributions of the target variables $\hat{a}_t$, $\hat{RTG}_t$, and $\hat{CTG}_t$. This allows the actor to infer feasible actions for given goals, and the critic to predict future goals under renewable uncertainty. The network parameters $\theta$ (actor) and $\theta'$ (critic) are optimized by minimizing a negative log-likelihood (NLL) loss:

\begin{equation}
\begin{aligned}
J_{\pi}(\theta) 
&= -\frac{1}{K+1} 
  \mathbb{E}_{\mathcal{B}_N} 
  \sum_{k=0}^{K}
  \Big[
  \log \pi_\theta \big(
  a_{t-k} \mid 
  \{ D \cup 
  \{ CTG_i, RTG_i, s_i, a_i \}_{t-K}^{t-1-k} \\
&\qquad\qquad\qquad\quad
  \cup 
  \{ CTG_{t-k}, RTG_{t-k}, s_{t-k} \} 
  \}
  \big)
  \Big]
\end{aligned}
\label{nll-actor}
\end{equation}

\begin{equation}
\begin{aligned}
J_{Q}(\theta') 
&= -\frac{1}{K+1} 
  \mathbb{E}_{\mathcal{B}_N} 
  \sum_{k=0}^{K}
  \Big[
  \log Q_{\theta'} \big(
  RTG_{t-k} \mid 
  \{ D, E \} \cup 
  \{ s_i, a_i, c_i, r_i \}_{t-K}^{t-k}
  \big)
  \\
&\quad + 
  \log Q_{\theta'} \big(
  CTG_{t-k} \mid 
  \{ D, E \} \cup 
  \{ s_i, a_i, c_i, r_i \}_{t-K}^{t-k}
  \big)
  \Big]
\end{aligned}
\label{nll-critic}
\end{equation}
where $\mathcal{B}_N \subset \mathcal{D}{\text{oracle}}$ is a mini-batch of $N$ samples drawn from the oracle dataset, and $K$ is the sequence length. Details of the network architecture and hyperparameters are provided in Table \ref{tab:hyper-all}.

\cref{figS6:actor-critic-loss} illustrates the NLL loss curves for the actor and critic, trained on the Dunkirk region–oriented oracle dataset. (This oracle dataset was generated by the scenario generator $G$, which was trained on Dunkirk’s historical wind-speed data.) To evaluate the effect of conditional tokens $D$ (design setting) and $E$ (renewable trend), we trained each network under different combinations of these tokens. In \cref{figS6:actor-critic-loss} (a), incorporating the design token $D$ in the actor network yields a lower NLL loss for action $a_t$, improving the accuracy of the actor’s inferred action distribution under varying design settings. However, adding the renewable-trend token $E$ provides no noticeable improvement for the actor. By contrast, \cref{figS6:actor-critic-loss} (b) and (c) show that including both $D$ and $E$ tokens in the critic significantly reduces the NLL losses for the $RTG_t$ and $CTG_t$ predictions. This suggests that predictions of $RTG_t$ and $CTG_t$ depend strongly on both the design setting and the renewable trend. These results serve as a preliminary ablation study on the conditional tokens, indicating that the optimal configuration is to use token $D$ for the actor $\pi_{\theta}$, while using both $D$ and $E$ for the critic $Q_{\theta'}$ (Further details ablation studies and quantitative analysis will be described in \cref{sec3.4}.)

\noindent\textbf{Offline operation} Given fixed design token $D$, the offline operation assumes the renewable trend token $E$ is fixed during system operation. The token $E$ can be obtained in two ways: either from historical renewable data reflecting seasonal trends, or from synthetic scenarios generated by the trained scenario generator $G_{\theta}$. The former approach supports region-specific operations across different seasons, while the latter enables parallelized operational problem solving under given design and synthetic scenario set for uncertainty quantification or second-stage problem solving in co-optimization loop. 

Given  $E$ and $D$, the offline operation proceeds as follows:
\begin{enumerate}
    \item \textbf{Sample candidate actions:} The actor $\pi_\theta$ samples $N$ candidate actions $\{\hat{a}_t^i\}_{i=1}^N$ from its inferred distribution, conditioned on the current trajectory, and the design token $D$.

    \item \textbf{Evaluate action:} For each candidate action $\hat{a}_t^i$, compute the resulting carbon emission $c_t^i$ and profit $r_t^i$ at time $t$

    \item \textbf{Predict long-term goals:} For each candidate tuples $(s_t, \hat{a}_t^i, c_t^i, r_t^i)$, the critic $Q_{\theta'}$ predicts the long-term reward and carbon emission $(\hat{RTG}_t^i, \hat{CTG}_t^i)$. 

    \item \textbf{Select the optimal action:} The action maximizing the predicted return while satisfying the carbon constraint is selected:
    \[
    a_t^* = \arg\max_{i=1,\dots,N} \ \hat{RTG}_t^{(i)} \quad \text{s.t.} \quad \hat{CTG}_t^{(i)} \leq 0
    \]

    \item \textbf{Update next goals:} Apply the selected action $a_t^*$ to the system. Then update the goal tokens for the next time step:
    \[
    \hat{RTG}_{t+1} = RTG_t^* - r_t^*, \quad \hat{CTG}_{t+1} = CTG_t^* - c_t^*
    \]
\end{enumerate}
\cref{algo:offline} provides a summarized pseudocode of this offline operation procedure for the proposed actor–critic model.

\noindent\textbf{Online operation} In the online setting, the renewable trend token $E$ is not known a prior since we do not have access to a full-horizon renewable scenario in advance. Instead, the $E$ must be inferred dynamically from recent renewable observations in real time. In this study, we proposed a forecasting-based policy update algorithm (inspired by the approach of Chen et al., 2018 \cite{chen2018unsuperviseddeeplearningapproach}) in which the noise vector $z$ (input to the trained generator $G_{\theta}$) is optimized to produce future renewable scenario set that align with the latest observations. This allows the actor–critic model to adapt in real time by estimating $E$ dynamically, without requiring complete historical or future renewable information. Thus, the proposed algorithm supports real-time operation under an optimized system design obtained via co-optimization.

Given fixed design token $D$ and vector of observed renewable $w_{obs} = [w_{1}\cdots w_t ]^T$ up to current time $t$, the online operation proceeds as follows:

\begin{enumerate}
\item \textbf{Sampled candidate tuples:} The actor $\pi_\theta$ samples $N$ candidate tuples $\{(s_t,\hat{a}_t^i,c_t^i,r_t^i)\}_{i=1}^N$ following the same procedure as in offline operation step 1 and 2. 

\item \textbf{Infer the renewable trend token $E$:} To estimate the $E$ under uncertainty, sample a $M$ batch noise $\{z_j\}_{j=1}^M$ from a multivariate Gaussian distribution. Map each $z_j$ through $G(z_j)$ to produce a predicted renewable scenario $w_{\text{pred}}^j = \{w_t^{(r,j)}\}_{t=0}^{T-1}$ of length $T$. The $\{z_j\}_{j=1}^M$ are optimized to minimize the error between the $w_{\text{obs}}$ and the corresponding $t$ segment of $w_{\text{pred}}^j$. An adversarial loss term $-D(G(z_j))$ is added to the optimization to ensure the predicted scenarios remain realistic. After optimization, use each predicted scenario $w_{\text{pred}}^j$ to update the corresponding renewable trend token $E_j$

\item \textbf{Propagate uncertainty with critic:} Given $\{(s_t, \hat{a}_t^i, c_t^i, r_t^i)\}_{i=1}^N$ and $\{E_j\}_{j=1}^M$, the critic $Q{\theta'}$ evaluates each action candidate under inferred renewable trend set. It infers the distribution of the goal tokens for each pair $(\hat{a}_t^i, \{E_j\}_{j=1}^M)$, producing updated samples of $\hat{RTG}_t$ and $\hat{CTG}_t$ that incorporate the uncertainty in renewable trends.

\item \textbf{Select action and update:} Perform the action selection and goal token update as in offline operation steps 4–5.

\end{enumerate}
\cref{algo:online} provides a pseudocode of this online operation procedure for the proposed actor-critic model.

\subsection{Experiment}\label{sec3.4}
This section evaluates both the computational efficiency and the operational performance of the proposed actor–critic agent for generalized power-to-methanol operation. We begin by outlining the experimental setup, including the validation dataset, deterministic LP benchmark, the RL-based comparison model. We then present empirical results that demonstrate: (i) improved computational efficiency compared with the LP solver when solving large sets of scenarios in parallel, and (ii) superior operational performance—higher profit, improved carbon-emission constraint satisfaction, and more accurate long-term goal predictions $(RTG, CTG)$—in both offline and online operation settings under varying design conditions and renewable scenarios.

\subsubsection{Experiment setup}\label{sec3.4.1}
\noindent\textbf{Dataset}
For benchmark comparison, we used the 2023–2024 year historical renewable-power and grid-price validation dataset for the Dunkirk region (details in \cref{tab:region-lat-long}), which is unseen by both the generator $G$ and the actor-critic agent models. This provides an unbiased evaluation because all models were trained exclusively on the synthetic oracle dataset constructed from the 2015–2022 historical data.

To assess generalizability across different design configurations, four representative design categories were defined from the full design space in \cref{tab:design_space}. The percentage ranges specify the portion of each variable’s design range from which samples were drawn:
\begin{itemize}
    \item \textbf{Base}: middle 40--60\% of all design-variable ranges.
    \item \textbf{Buffered}: upper 40--80\% for storage units ($C_\text{CHT}, C_\text{BESS}$) and lower 0--20\% for production units ($\alpha_\text{PEM}, \dot{m}_{\text{MeOH}}$).
    \item \textbf{Responsive}: lower 0--20\% for storage units and middle 40--60\% for production units.
    \item \textbf{Max-product}: middle 40--60\% for storage units and upper 60--100\% for production units.
\end{itemize}

For offline evaluation, 25 designs were sampled using Latin hypercube sampling \cite{loh1996latin} within each category (100 total) and combined with 100 set of validattion scenarios of renewable power and grid price, resulting in $4 \times 25 \times 100$ experiment instances. For online evaluation, where inference of the renewable-trend token $E$ increases computational cost, 250 designs per category (1,000 total) were paired with 1,000 validation scenarios set. s.

\noindent\textbf{Benchmark} We compared our proposed actor-critic model with the following four representative RL-based model:
\begin{itemize}
    \item \textbf{DRL}: A deep reinforcement learning agent trained with Proximal Policy Optimization (PPO) \cite{schulman2017proximal}, representing a state-of-the-art online RL method for continuous action space. The design vector is provided as an additional input to condition the policy network across different design settings. Due to the strong sensitivty in selection of hyperparameter, especially under varying design configurations, we applied population-based hyperparameter optimization (PBT) \cite{jaderberg2017population} to ensure competitive performance.
    
    \item \textbf{BC}: A behavior cloning model \cite{torabi2018behavioral} sharing the same network architecture as DRL, but trained purely via supervised learning on the state–action pairs $(s, a)$ in the oracle dataset  $\mathcal{D}_{\text{oracle}}$.
    
    \item \textbf{DRL+BC}: A hierarchical approach where a DRL agent refines the action parameters predicted by a pretrained BC model. The BC parameters are frozen, and the DRL component is trained via PPO to enable exploration while leveraging BC policy. Both components share the same neural architecture.
    
    \item \textbf{DT}: A decision transformer \cite{chen2021decision} conditioned on manually assigned initial goal values $(RTG_0, CTG_0)$, set to the average target values computed from the $\mathcal{D}_{\text{oracle}}$.
    
    \item \textbf{ST}: A vanilla actor-critic based Saformer \cite{zhang2023saformer} built upon DT without additional token embeddings of $D$ and $E$.   
\end{itemize}
The network architecture and hyperparameters are listed in \cref{tab:hyper-all}. In this study, actor-critic model architecture of ST and proposed PT share same architecture of DT except the additional embedding layers of $D$ and $E$ tokens. 
 
\subsubsection{Computational efficiency analysis}\label{sec3.4.2}
We first compare the computational cost of the proposed model against Gurobi \cite{gurobi}, a CPU-based MILP solver, over problem sets of different sizes solved in parallel. For a fair comparison, while the proposed model runs on a single GPU, Gurobi solves the LP problem set (\cref{sec1.2}) varying CPU cores from 1 to 16, with each process restricted to single-thread. All computations were performed on a computing server equipped with an Intel Xeon Platinum 8480C CPU (56 cores, 112 threads, up to 3.8 GHz, 2.01 TB RAM) and a single NVIDIA H200 GPU (140 GB memory, 700 W TDP). 

\cref{figS7:solution-time} shows the solution time across problem-set sizes ranging from 10 to 10,000. The Gurobi solver benefits from increasing CPU cores when solving large problem sets in parallel. For instance, compared with single core, 16 core achieved speed-ups of 0.38, 0.71, and 0.91 orders of magnitude for problem set sizes of 10, 100, and 10,000. Despite these gains in MILP solver, the proposed model still attains lower solution times for most problem sizes, and its gap increased as the problem sizes grows. For example, for a problem size of 10, the proposed model requires 4.2 secs whereas 16-core Gurobi requires 2.9 secs, primarily due to the base computational overhead of transformer-network operations. However, for problem sizes over 20, the proposed model consistently shows lower solution times than the Gurobi. This is because the proposed model relies on single-feedforward inference of optimal action and goals (\cref{algo:offline}) supported by highly parallelized GPU, whereas Gurobi must execute the full MILP procedure on CPU. Consequently, as the problem size increases from 100 to 10,000, the proposed model achieves speed-ups of 0.366 to 0.70 orders of magnitude, compared with MILP solver. For instance, for problem-set sizes of 1,000 and 10,000, the proposed model solves the problems in 17.6 and 160.7 secs, compared with 84.8 and 815.3 secs for the Gurobi solver. This improvement in solution time is critical for two stage scenario based design optimization, which requires solving large operational problem sets with scenario sets in limited computational times.
  
\subsubsection{Offline operation performance compared with the benchmark model}\label{sec3.4.3} 
The monthly operation problem defined in \cref{sec1.2} aims to maximize monthly profit while enforcing a negative carbon emission constraint. Accordingly, we evaluate the optimality gap in profit maximization and the constraint violation of negative carbon emission between dynamic operational policy models and the deterministic MILP solution, using the four design categories and validation dataset described in \cref{sec3.4.1}. For offline operation settings, renewable trend token $E$ is precomputed under given renewable scenario, and \cref{algo:offline} is applied to the PT critic models for goal ($RTG_t, CTG_t$) predictions. 

As shown in \cref{figS8:offline-performance}, the proposed PT model consistently outperforms all benchmark models across the four design categories, achieving the lowest optimality gap, the lowest constraint violation, and the smallest performance variation across diverse monthly scenarios. Furthermore, the progressive performance improvements from RL-based models to PT highlight how oracle-trajectory learning, the incorporation of critic model, and the embedding of design and trend tokens ($D, E$) collectively enhance generalized operational performance. For instance, in the base design case, BC and RL-based models (DRL, DRL+BC) yield an average optimality gap of 89.6\% with average constraint violation of 2565.6 ton/month, where only two solution satisfy the negative-emission constraint. In contrast, DT and ST models improve the optimality gap to 79.8\% and 42.4\%, respectively, supporting the advantage of learning from oracle trajectories rather than from one-step state–action pairs. Moreover, unlike DT that relies on manually assigned goals, the ST critic dynamically predicts goals $(RTG_t, CTG_t)$ and conditions the actor’s action distribution on these goals. The sampled action is then fed back to the critic with the resulting reward and cost to update the goal distribution. This closed-loop actor–critic structure prevents the operational policy from deviating from learned oracle trajectory and maintains consistency between the predicted goals and the realized state–action evolution.

Despite these improvements, ST still exhibits an average constraint violation of 844.9 ton/month and provides feasible solutions in only 47.81\% of scenarios across the four design categories. In contrast, the proposed PT model achieves both lower optimality gaps and lower constraint violations than ST across all scenarios. Specifically, in the base, responsive, and max-product design cases, the PT-based solutions capture more carbon than the MILP solution, yielding average 91.0, 0.2, and 80.9 ton/month higher carbon capture than that of the MILP solution, respectively. Since ST and PT share the same actor–critic architecture, these improvements are derived from PT critic model’s enhanced goal-prediction accuracy, enabled by the additional design and renewable trend token embeddings ($D, E$), which is consistent with the lower critic training losses shown in \cref{figS6:actor-critic-loss} (b) and (c). To further verify this, we computed the continuous ranked probability score (CRPS) for $RTG_t$ and $CTG_t$, which measures the probabilistic error of goal predictions. As shown in \cref{tab:crps}, incorporating $D$ and $E$ token to critic model improves long-term goal prediction across all design and scenario cases, and \cref{figS9:offline-goal-prediction} visualizes this enhanced goal-prediction accuracy during dynamic operation. In summary, improved goal inference allows the PT critic to generate more accurate goal trajectories and guide the actor accordingly, leading to superior performance in both profit maximization and satisfaction of the negative-emission constraint.

\subsubsection{Online operation performance compared with the benchmark model}\label{sec3.4.4}
In online operation settings, the renewable trend token $E$ is not provided and must be dynamically inferred during system operation. Therefore, except for the benchmark models that do not require the $E$ token for system operation, \cref{algo:online} is solely applied to the proposed PT model to dynamically forecast a potential set of $E$ tokens from the observed renewable data and infer potential action for long-term constraint satisfaction under renewable uncertainty. The benchmark model follows the same calculation procedure used in the offline operation setting.

\cref{figS10:online-performance} demonstrates progressive performance improvements from the RL-based model to the proposed PT model, consistent with the offline operation results (\cref{sec3.4.4}). For example, compared with the BC and RL-based models, the oracle-trajectory-learning-based critic models of both ST and PT effectively adjust the actor’s action distribution to satisfy long-term constraints during operation trajectory evolving. As a result, the ST and PT models achieve 76\% and 75\% feasible solutions, whereas the BC and RL-based models yield only 51\% feasible solutions under negative carbon emission constraints.

Although ST and PT exhibit similar levels of optimality gap and feasibility ratio, the constraint violation of the PT model is significantly lower at 28.1 ton/month, compared with 78.16 ton/month for the benchmark ST model across the four design categories. This improvement comes from the higher goal-prediction accuracy of the critic models, and the lower CRPS values for PT shown in \cref{tab:crps} and the example goal-prediction profiles in \cref{figS11:online-goal-prediction} further support this finding.

In summary, under online operation without access to historical renewable datasets, the results still show that the proposed PT with \cref{algo:online} successfully enables dynamic inference of the $E$ token, which is crucial information for accurate goal prediction, and maintains operational performance close to the deterministic global solution. It achieves the lowest optimality gap and more than 75\% constraint satisfaction, outperforming all other dynamic policy models.

\clearpage
\refstepcounter{section}
\section*{Supplementary Figures}\label{sec:figures}
\setcounter{figure}{0}
\renewcommand{\thefigure}{S\arabic{figure}}
\begin{figure}[h]
\centering
\includegraphics[width=1.0\textwidth]{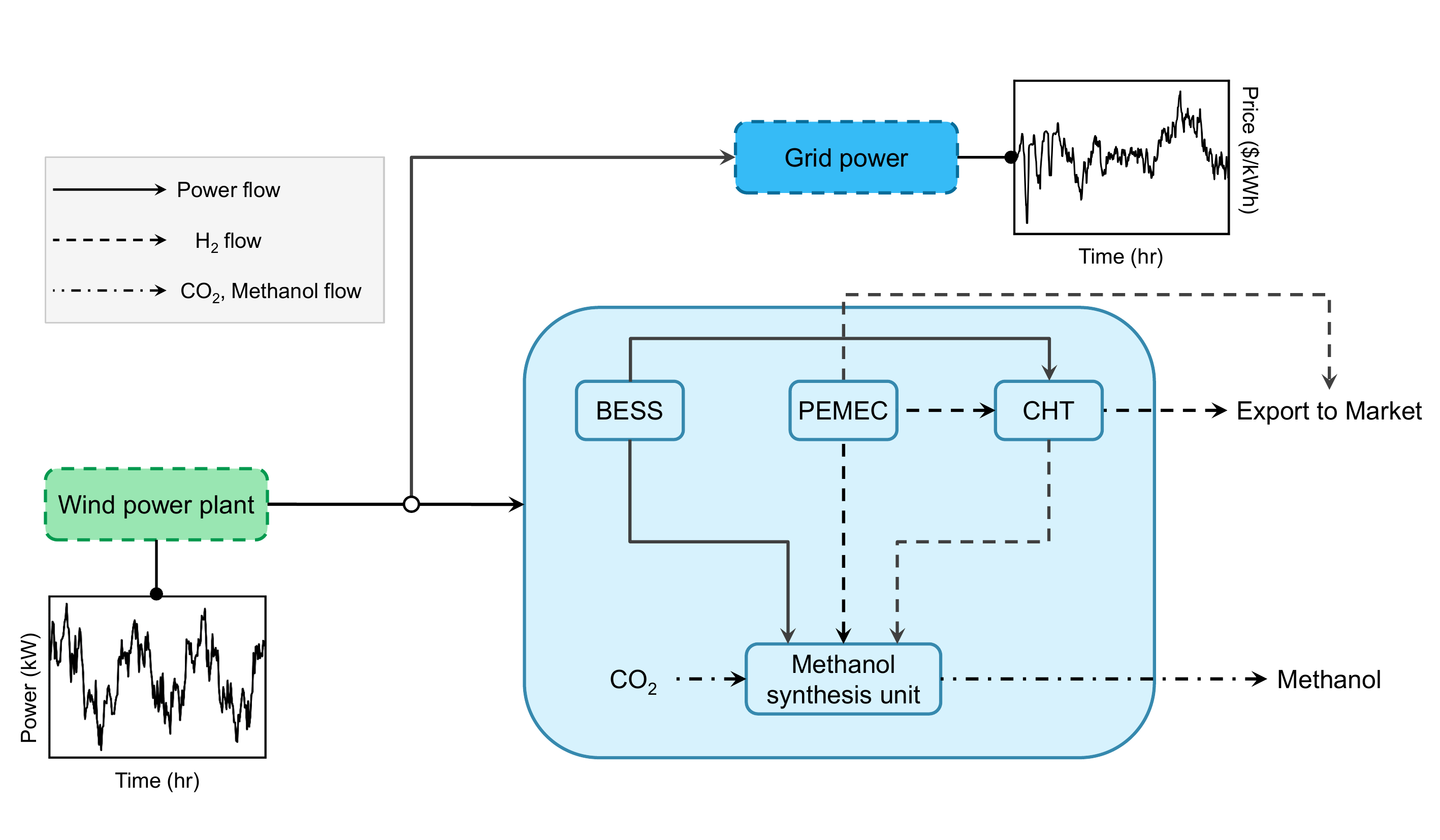}
\caption{Schematic of power-to-methanol system}\label{fig1:system-diagram}
\end{figure}

\begin{figure}[h]
\centering
\includegraphics[width=1.0\textwidth]{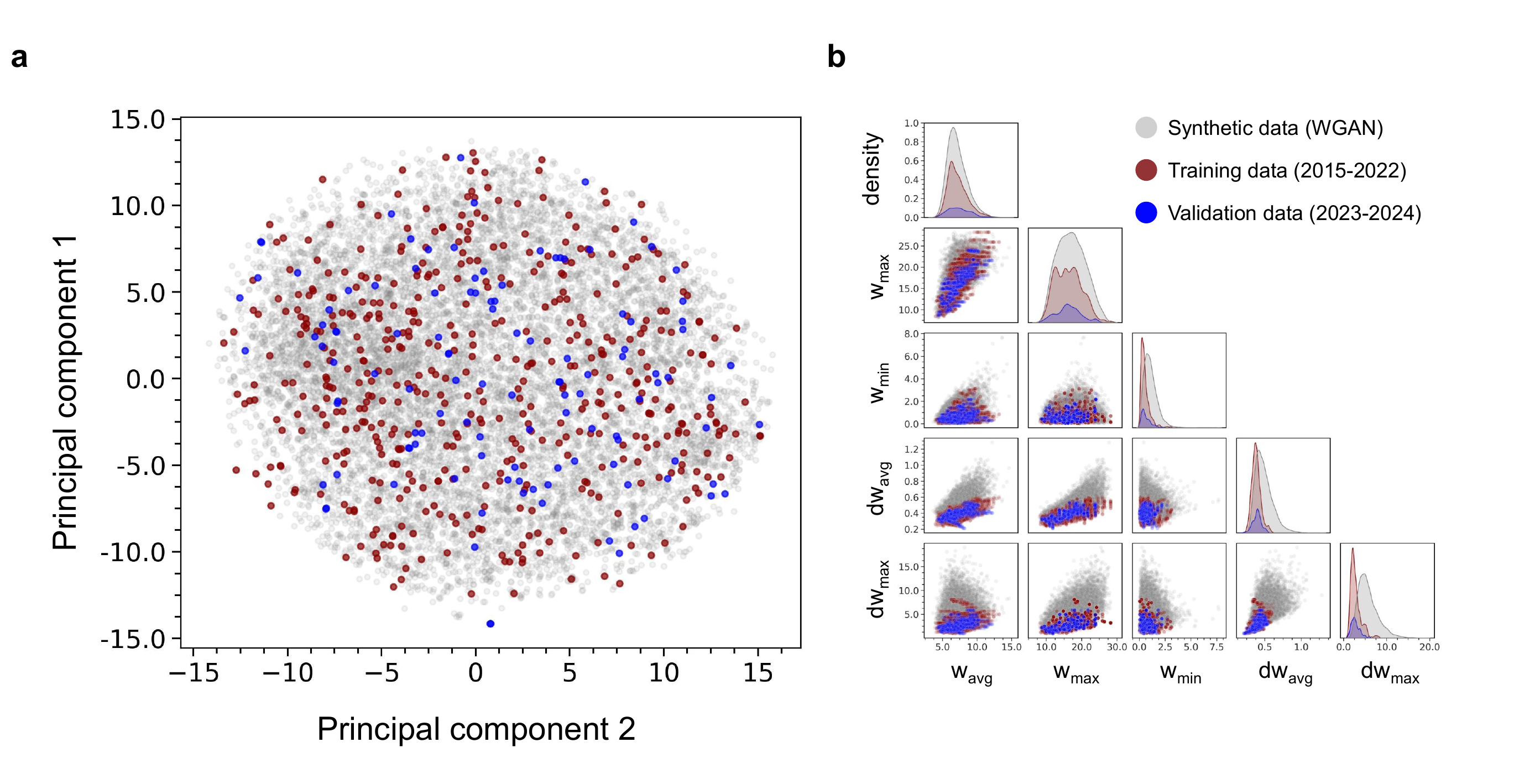}
\caption{Comparison of real and synthetic wind-speed data distribution in the Dunkirk (France) region: (a) t-SNE visualization and (b) key operational feature distributions}\label{fig2:tsne-dunkirk}
\end{figure}

\begin{figure}[h]
\centering
\includegraphics[width=1.0\textwidth]{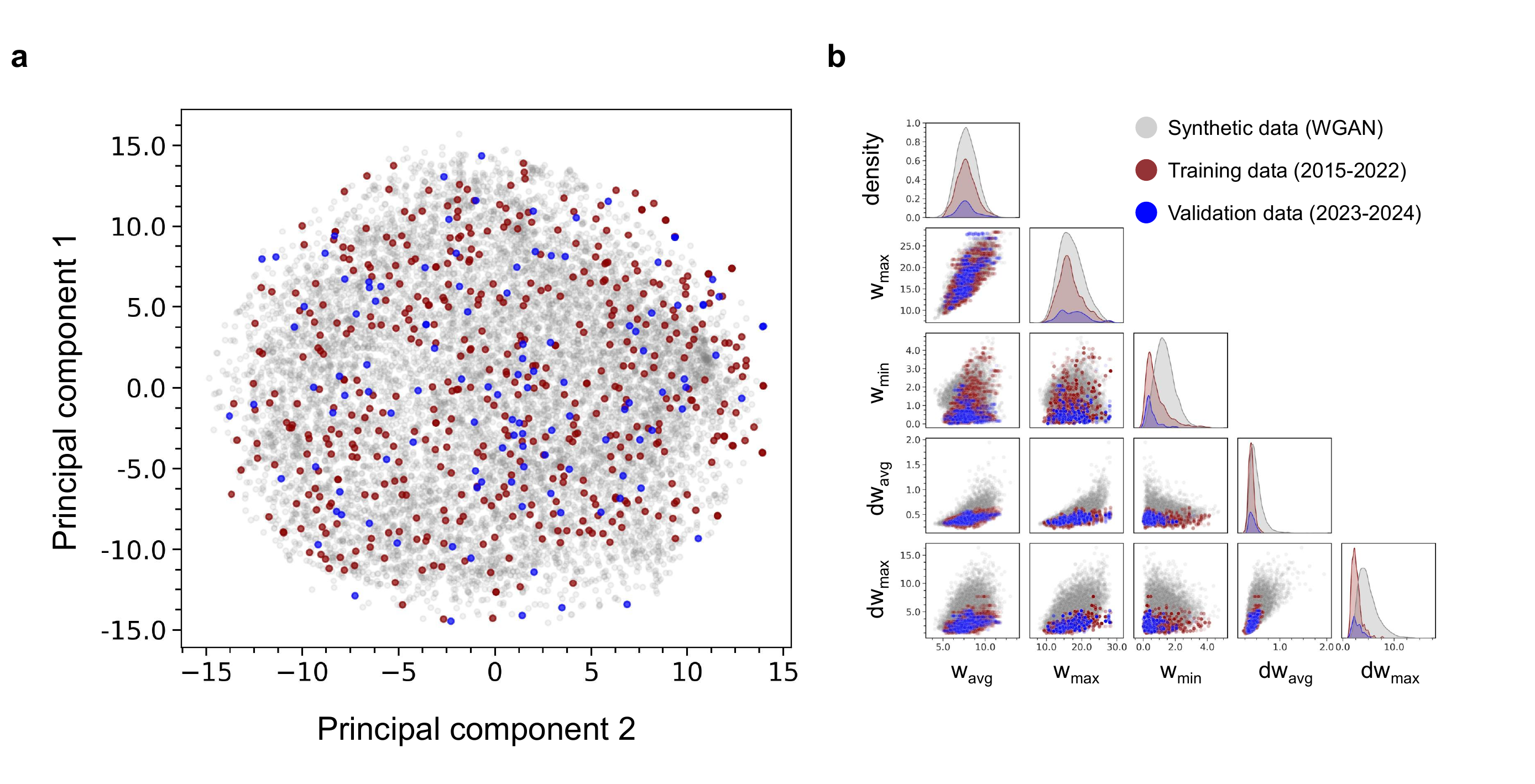}
\caption{Comparison of real and synthetic wind-speed data distribution in the Skive (Denmark) region: (a) t-SNE visualization and (b) key operational feature distributions}\label{fig3:tsne-skive}
\end{figure}

\begin{figure}[h]
\centering
\includegraphics[width=1.0\textwidth]{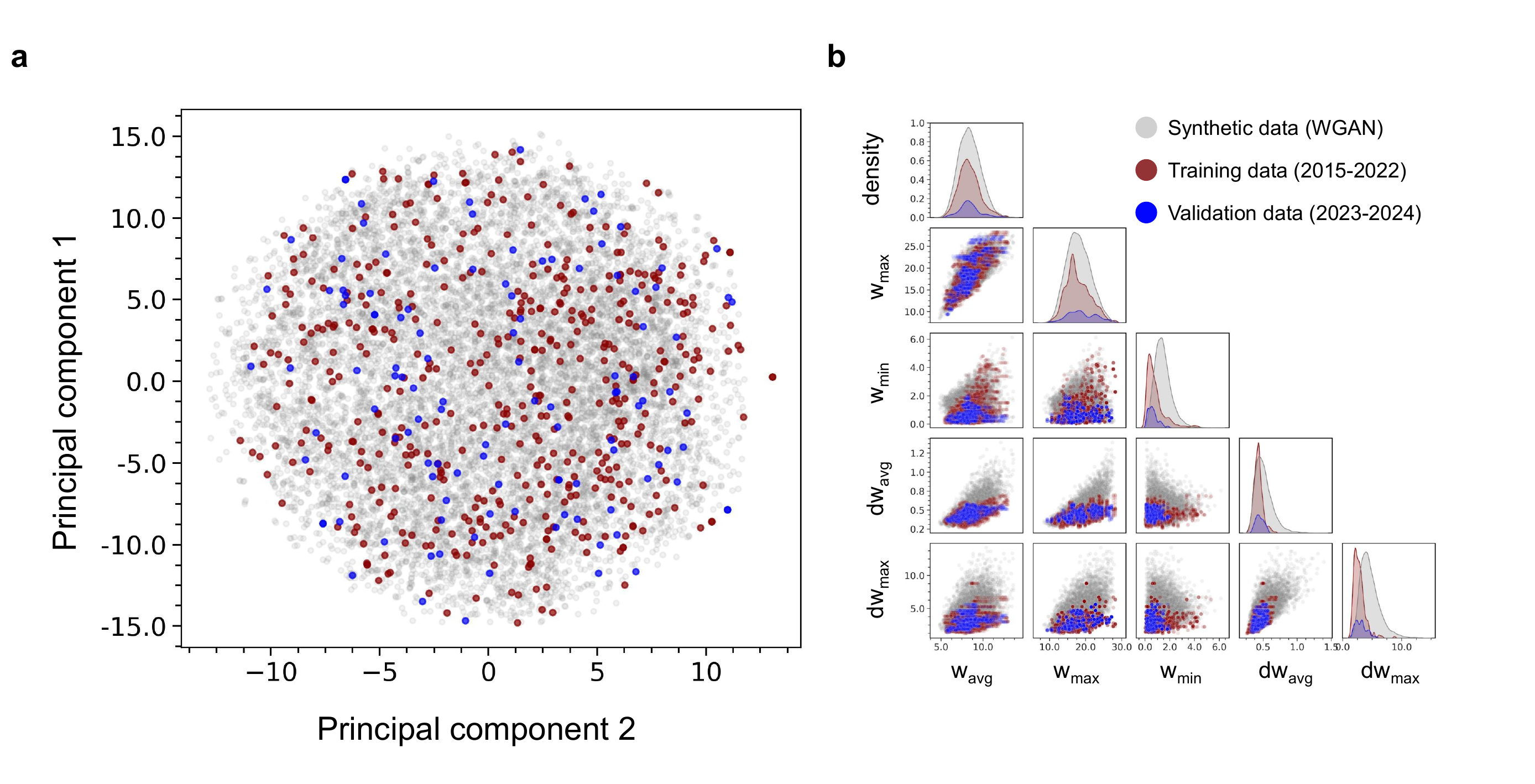}
\caption{Comparison of real and synthetic wind-speed data distribution in the Fredericia (Denmark) region: (a) t-SNE visualization and (b) key operational feature distributions}\label{fig4:tsne-fredericia}
\end{figure}

\begin{figure}[h]
\centering
\includegraphics[width=1.0\textwidth]{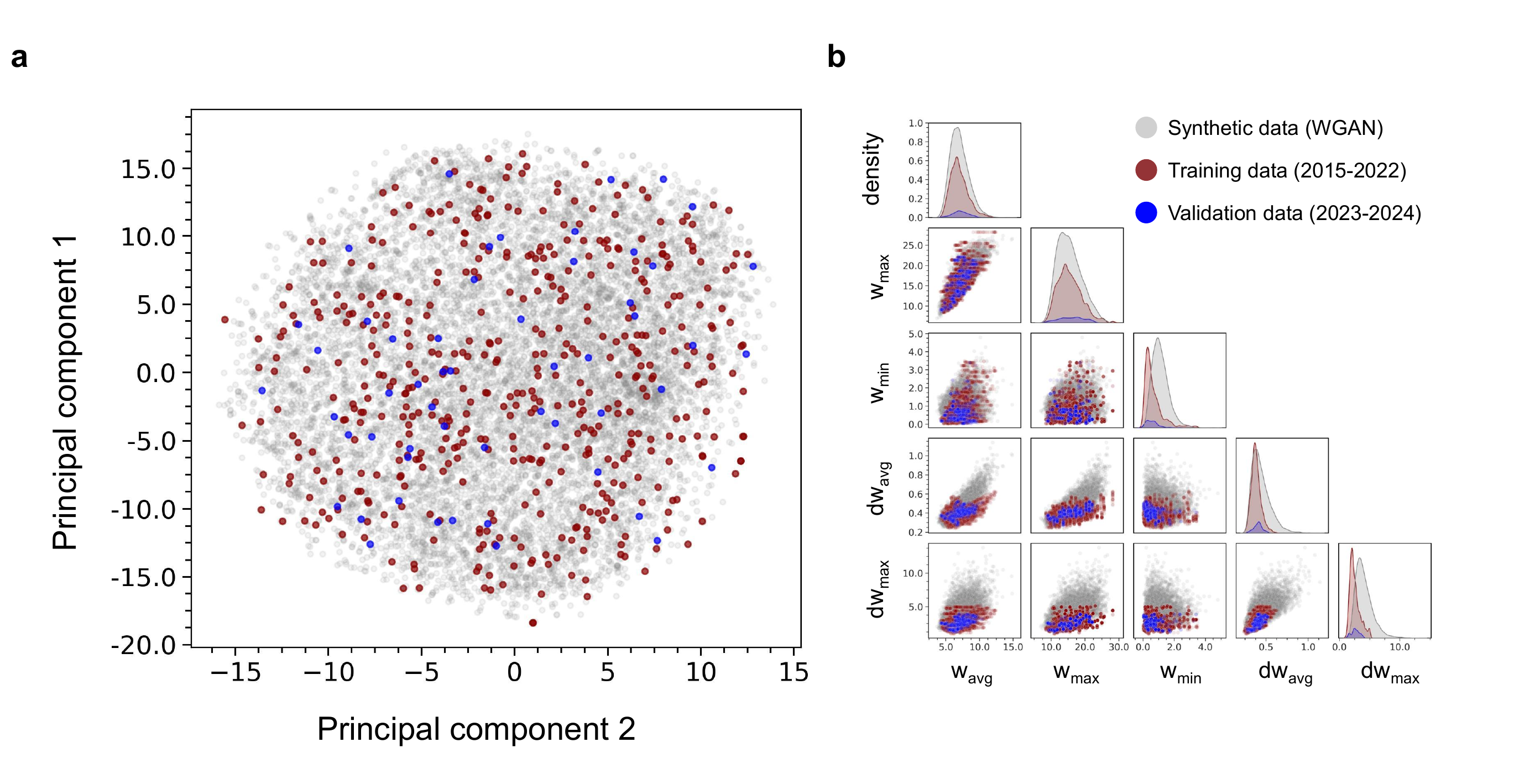}
\caption{Comparison of real and synthetic wind-speed data distribution in the Weener (Germany) region: (a) t-SNE visualization and (b) key operational feature distributions}\label{fig5:tsne-weener}
\end{figure}

\begin{figure}[h]
\centering
\includegraphics[width=0.72\textwidth]{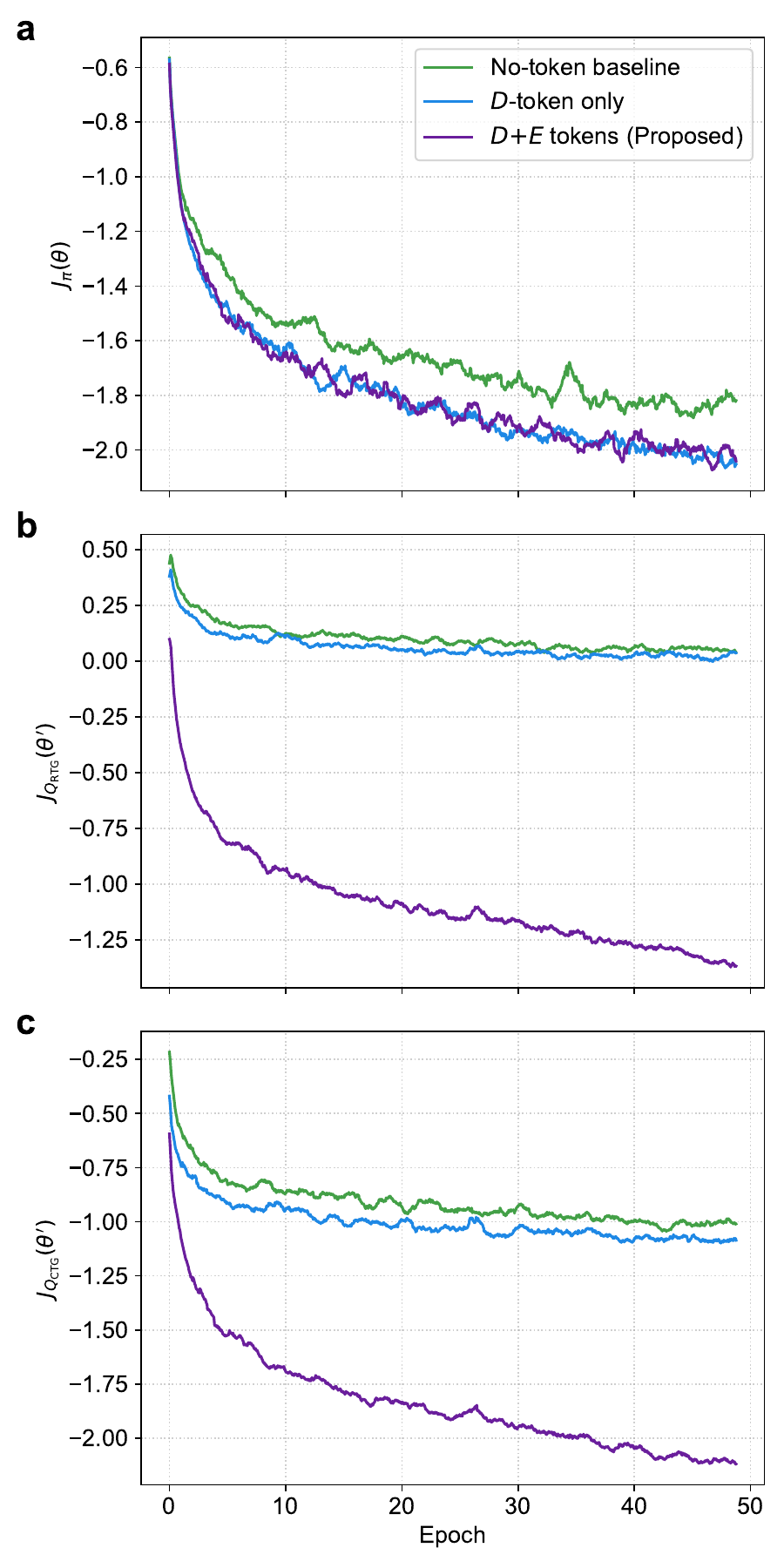}
\caption{
Training losses of the actor and critic networks under various combinations of design ($D$) and environment ($E$) tokens. 
(a) Actor loss $J_{\pi}(\theta)$. 
(b,c) Separated critic losses of $RTG_t$ and $CTG_t$.
}
\label{figS6:actor-critic-loss}
\end{figure}

\begin{figure}[h]
\centering
\includegraphics[width=1.0\textwidth]{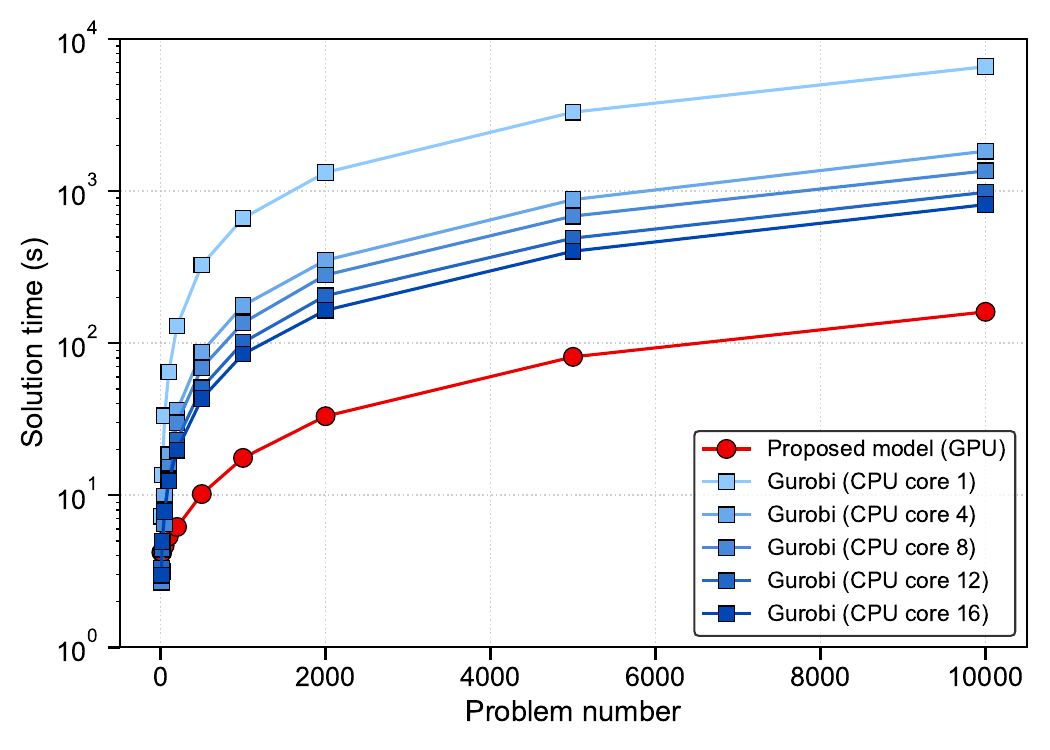}
\caption{Solution time of the proposed model and the Gurobi benchmark across different problem sizes}
\label{figS7:solution-time}
\end{figure}

\begin{figure}[h]
\centering
\includegraphics[width=1.0\textwidth]{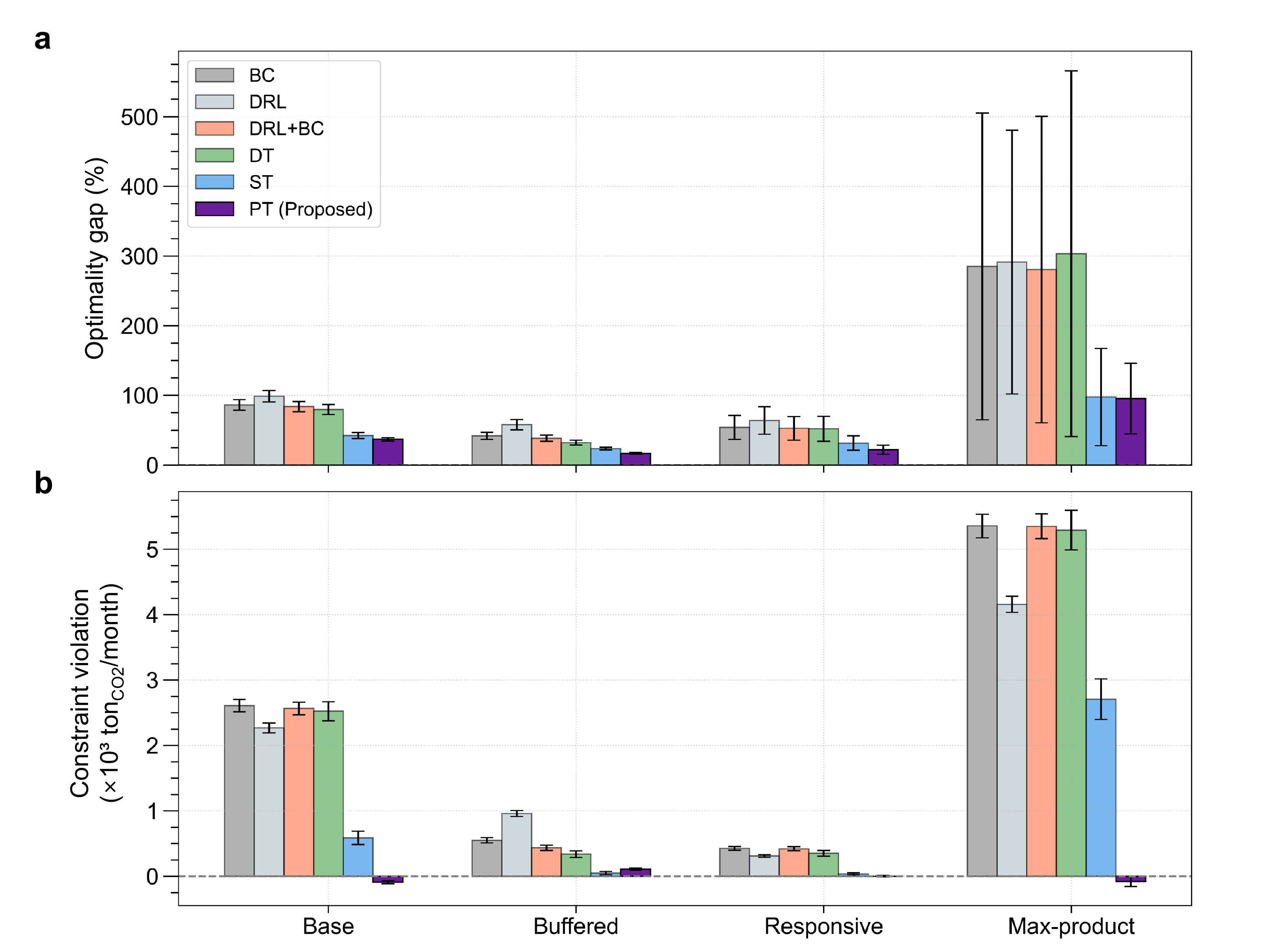}
\caption{Offline operation performance comparison of operational policy models across different design categories: (a) optimality gap from MILP solver (Gurobi), and (b) constraint violation of negative carbon emission. \\
\footnotesize *Error bars represent 10\% of the standard deviation}
\label{figS8:offline-performance}
\end{figure}

\begin{figure}[h]
\centering
\includegraphics[width=1.0\textwidth]{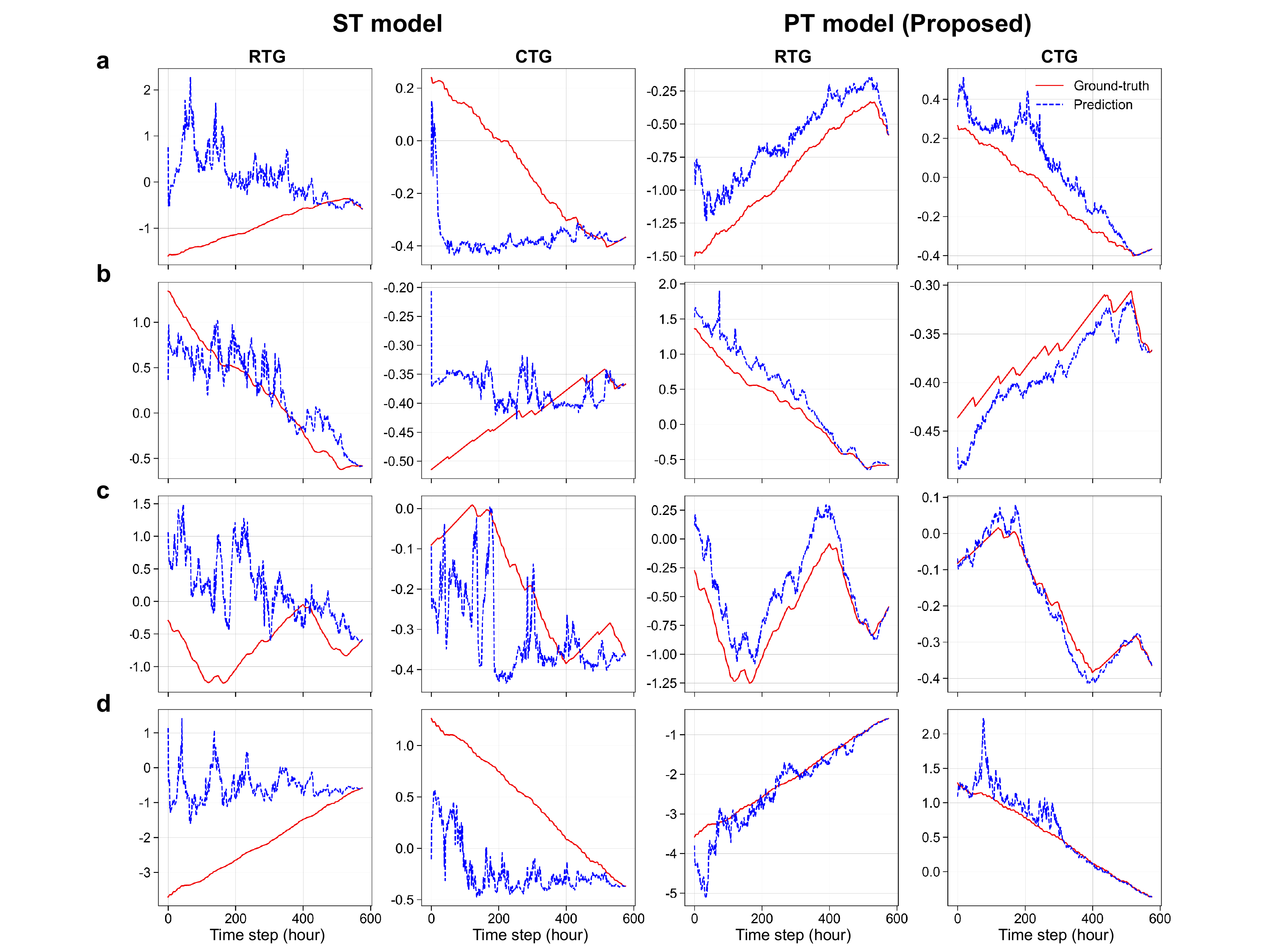}
\caption{Comparison of goal prediction performance of the ST and PT critic models in offline operation settings across the four design categories: (a) Base, (b) Buffered, (c) Responsive, and (d) Max-product.}
\label{figS9:offline-goal-prediction}
\end{figure}

\begin{figure}[h]
\centering
\includegraphics[width=1.0\textwidth]{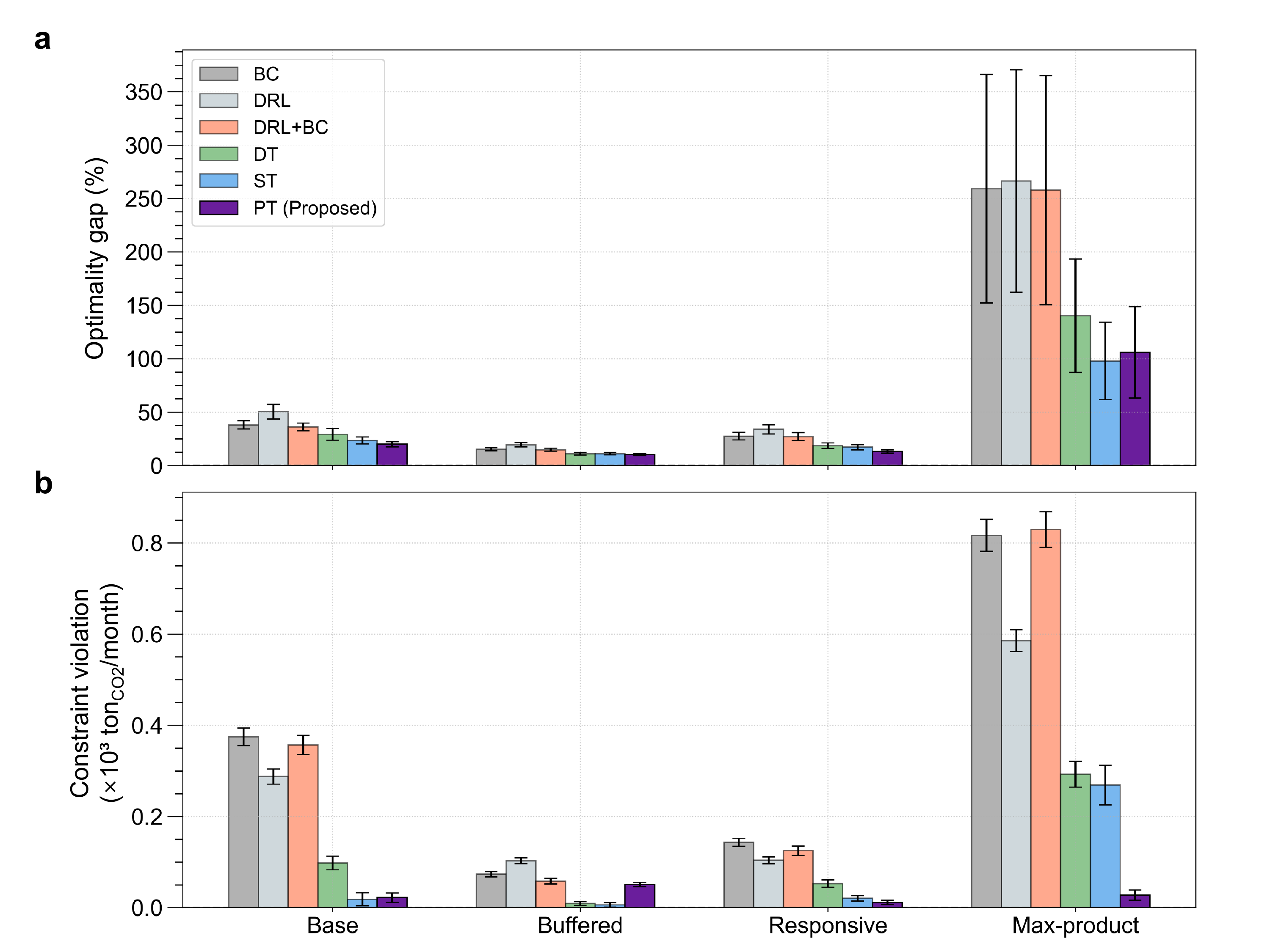}
\caption{Online operation performance comparison of operational policy models across different design categories: (a) optimality gap from MILP solver (Gurobi), and (b) constraint violation of negative carbon emission. \\
\footnotesize *Error bars represent 10\% of the standard deviation}
\label{figS10:online-performance}
\end{figure}

\begin{figure}[h]
\centering
\includegraphics[width=1.0\textwidth]{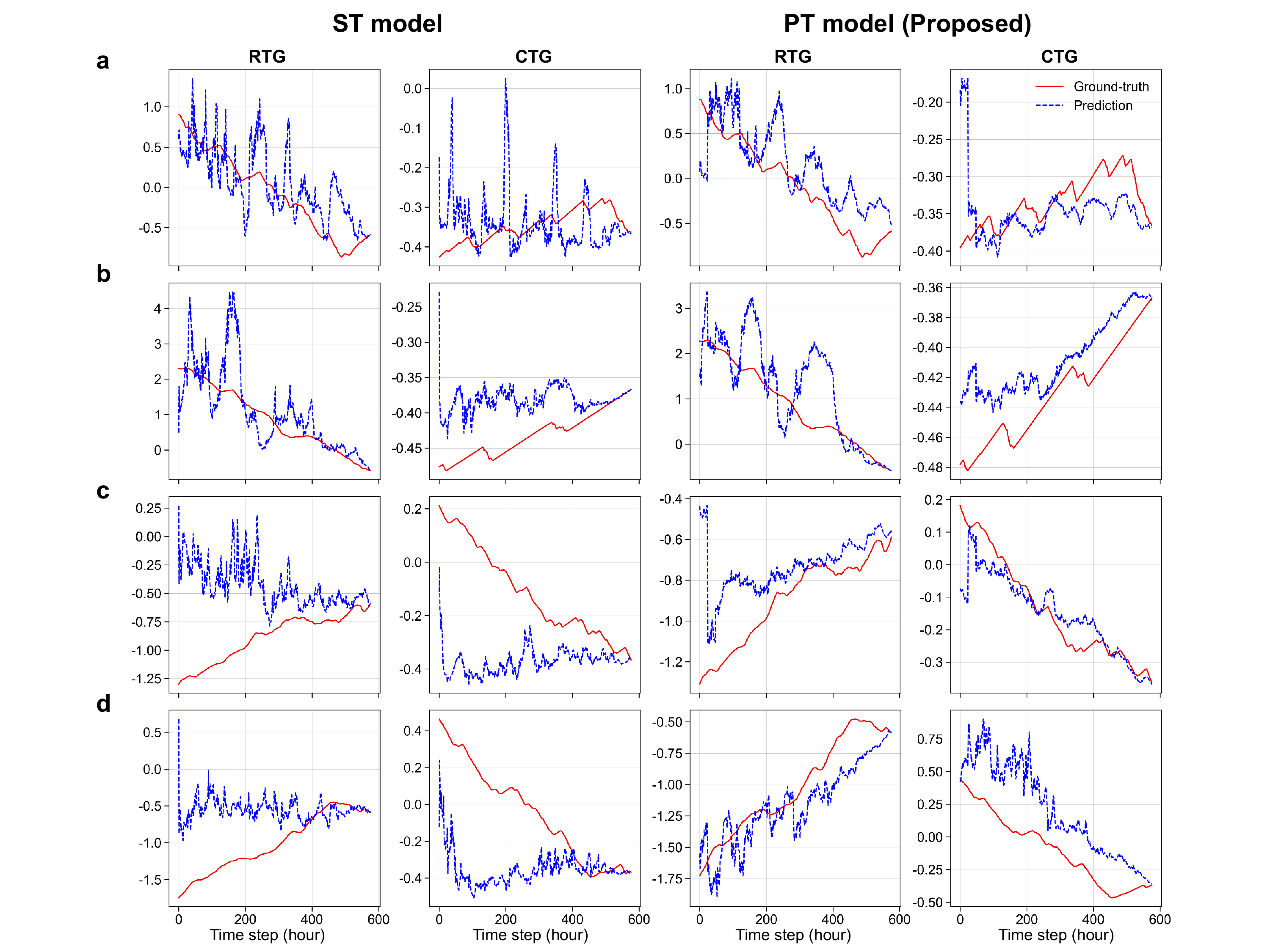}
\caption{Comparison of goal prediction performance of the ST and PT critic models in online operation settings across the four design categories: (a) Base, (b) Buffered, (c) Responsive, and (d) Max-product.}
\label{figS11:online-goal-prediction}
\end{figure}

\begin{figure}[h]
\centering
\includegraphics[width=1.0\textwidth]{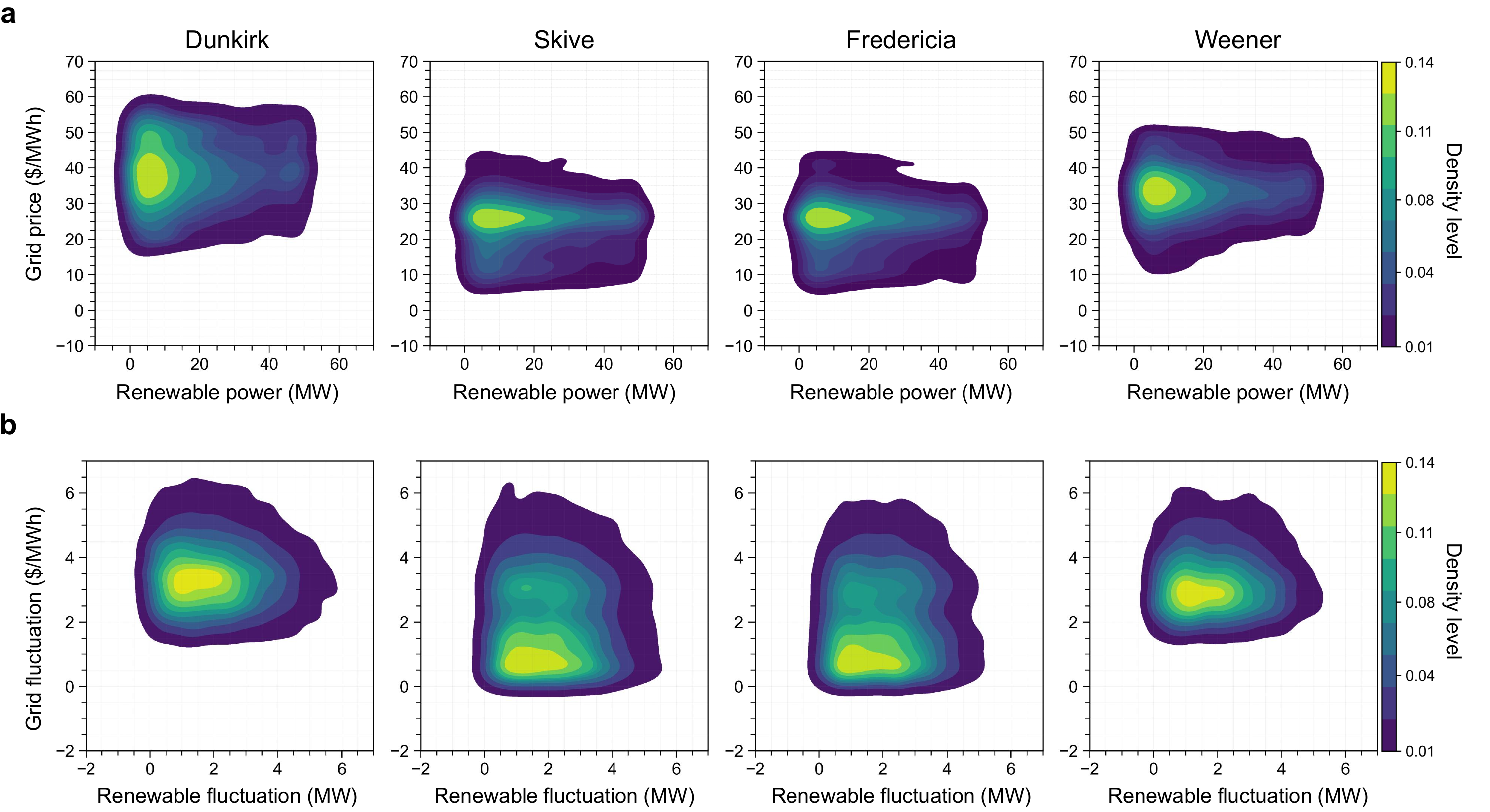}
\caption{Distributions of grid price and renewable power across the four target regions: (a) Distribution of daily-averaged grid price and renewable power, (b) Distribution of daily-averaged fluctuations in grid price and renewable power.}
\label{figS12:renew-grid-distributon}
\end{figure}

\begin{figure}[h]
\centering
\includegraphics[width=1.0\textwidth]{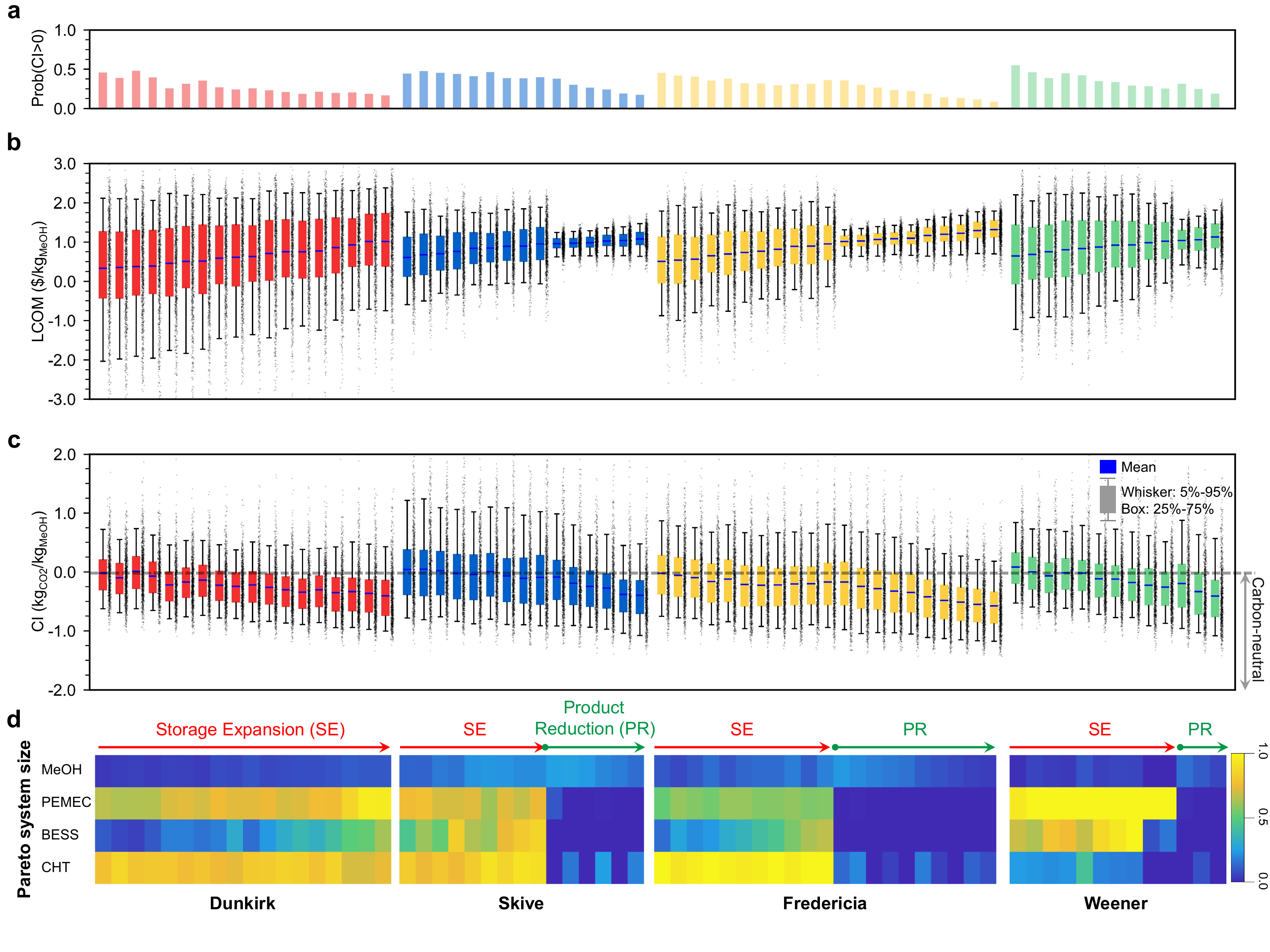}
\caption{Performance comparison of Pareto-optimal power-to-methanol designs across the target regions: (a) Probability of net-positive carbon emissions (CI$>0$), (b) Distribution of levelized cost of methanol production (LCOM), (c) Distribution of carbon intensity (CI), (d) Normalized values of the Pareto-optimal system designs relative to their respective design space in \cref{tab:design_space}.}
\label{figS13:uq-pareto-result}
\end{figure}

\clearpage
\refstepcounter{section}
\section*{Supplementary Tables}\label{sec:tables}
\renewcommand{\thetable}{S\arabic{table}}
\setcounter{table}{0}

\begin{table*}[h]
\centering
\caption{Technical parameters of the Power-to-Methanol system}
\label{tab:technicalparameter}
\resizebox{\textwidth}{!}{
\begin{tabular}{lccccl}
\toprule
\textbf{Unit process} & \textbf{Parameter} & \textbf{Value} & \textbf{Unit} & \textbf{Reference} \\
\midrule
\multirow{3}{*}{BESS} 
 & $\eta_{C}$, $\eta_{D}$ & 0.95 & -- & \cite{kang2023optimal} \\ 
 & $\alpha$ & 0.3 & -- & \cite{kang2023optimal} \\
 & $\eta_l$ & $6.94\times10^{-5}$ & -- & \cite{kang2023optimal} \\
 \midrule
 \multirow{2}{*}{PEMEC and CHT}
 & $SP_{H_2}$ & 55.7 & kW/(kg$_{H_2}$·h) & \cite{chen2021power} \\
 & $SPC_{H_2}$ & 3.003 & kW/(kg$_{H_2}$·h) & \cite{chen2021power} \\
 \midrule
 \multirow{3}{*}{Methanol synthesis unit}
 & $\gamma_{H}$ & 0.196 & kg$_{H_2}$/kg$_{\mathrm{MeOH}}$ & \cite{vo2022design} \\
 & $\gamma_{CO_2}$ & 1.436 & kg$_{CO_2}$/kg$_{\mathrm{MeOH}}$ & \cite{vo2022design} \\
 & $SP_{\mathrm{MeOH}}$ & 0.657 & kW/(kg$_{\mathrm{MeOH}}$·h) & \cite{vo2022design} \\
\bottomrule
\end{tabular}}
\end{table*}

\begin{table*}[h!]
\centering
\caption{Carbon tax of grid electricity in different regions \cite{taxfoundation_carbon_taxes_europe_2024}}
\label{tab:carbon-tax}
\begin{adjustbox}{max width=\textwidth}
\begin{tabular}{l c}
\toprule
\textbf{Region} & \textbf{Carbon tax ($\$/\mathrm{ton CO_2}$)} \\
\midrule
Dunkirk    & 47.96  \\
Skive      & 28.10  \\
Fredericia & 28.10  \\
Weener     & 48.39  \\
\bottomrule
\end{tabular}
\end{adjustbox}
\end{table*}

\begin{table*}[h!]
\centering
\caption{Hyperparameters used for WGAN-GP model training.}
\label{tab:gan-hyperparameter}
\begin{adjustbox}{max width=\textwidth}
\begin{tabular}{l c}
\toprule
\textbf{Parameter} & \textbf{Value} \\
\midrule
Learning rate & 0.0001 \\
Gradient penalty weight $\lambda$ & 10 \\
Batch size & 64 \\
Epoch & 15{,}000 \\
Bandwidth set $\sigma$ & $\{10, 15, 20, 50\}$ \\
\bottomrule
\end{tabular}
\end{adjustbox}
\end{table*}

\begin{table*}[h!]
\centering
\caption{The network structure of WGAN-GP}
\label{tab:gan-network}
\begin{adjustbox}{max width=\textwidth}
\begin{tabular}{l c c}
\toprule
\multicolumn{3}{c}{\textbf{Generator $G$}} \\
\midrule
\textbf{Layer} & \textbf{Filter size \textbar{} Stride} & \textbf{Activation size} \\
\midrule
Input & -- & 205 \\
Linear, BN, ReLU & -- & 1024 \\
Linear, BN, ReLU & -- & $512 \times 3 \times 3$ \\
ConvTrans, BN, ReLU & $512 \times 2 \times 2 \mid 2$ & $256 \times 6 \times 6$ \\
ConvTrans, BN, ReLU & $256 \times 2 \times 2 \mid 2$ & $128 \times 12 \times 12$ \\
ConvTrans, BN, Sigmoid & $1 \times 2 \times 2 \mid 2$ & $1 \times 24 \times 24$ \\
\midrule
\multicolumn{3}{c}{\textbf{Discriminator $D$}} \\
\midrule
\textbf{Layer} & \textbf{Filter size \textbar{} Stride} & \textbf{Activation size} \\
\midrule
Input & -- & $1 \times 24 \times 24$ \\
Conv, LeakyReLU, Dropout & $64 \times 2 \times 2 \mid 1$ & $64 \times 23 \times 23$ \\
Conv, LeakyReLU, Dropout & $128 \times 2 \times 2 \mid 2$ & $128 \times 11 \times 11$ \\
Conv, LeakyReLU, Dropout & $265 \times 2 \times 2 \mid 2$ & $256 \times 5 \times 5$ \\
Linear, LeakyReLU & -- & 1024 \\
Linear, LeakyReLU & -- & 128 \\
Linear & -- & 1 \\
\bottomrule
\end{tabular}
\end{adjustbox}
\vspace{1mm}
\parbox{0.9\textwidth}{\raggedright\footnotesize
* Conv: convolution layer, ConvTrans: convolutional transpose layer, BN: batch normalization.
}
\end{table*}

\begin{table*}[h!]
\centering
\caption{Wind speed dataset information}
\label{tab:region-lat-long}
\begin{adjustbox}{max width=\textwidth}
\begin{tabular}{l c c}
\toprule
\textbf{Region} & \textbf{Latitude range} & \textbf{Longitude range} \\
\midrule
Dunkirk    & [50.78$^{\circ}$, 51.23$^{\circ}$] & [2.08$^{\circ}$, 2.53$^{\circ}$] \\
Skive      & [56.46$^{\circ}$, 56.66$^{\circ}$] & [8.97$^{\circ}$, 9.07$^{\circ}$] \\
Fredericia & [55.46$^{\circ}$, 55.66$^{\circ}$] & [9.64$^{\circ}$, 9.79$^{\circ}$] \\
Weener     & [53.06$^{\circ}$, 53.26$^{\circ}$] & [7.25$^{\circ}$, 7.40$^{\circ}$] \\
\bottomrule
\end{tabular}
\end{adjustbox}
\vspace{1mm}
\parbox{0.9\textwidth}{\raggedright\footnotesize
* Wind speed at 50 m is collected from NASA POWER API at a spatial resolution of 0.05$^{\circ}$.
}
\end{table*}

\begin{table*}[h!]
\centering
\caption{Discriminative score of WGAN-generated renewable data}
\label{tab:discriminative-score}
\begin{adjustbox}{max width=\textwidth}
\begin{tabular}{l c}
\toprule
\textbf{Region} & \textbf{Discriminative score} \\
\midrule
Dunkirk    & $0.05 \pm 0.09$ \\
Skive      & $0.04 \pm 0.08$ \\
Fredericia & $0.01 \pm 0.08$ \\
Weener     & $0.01 \pm 0.09$ \\
\bottomrule
\end{tabular}
\end{adjustbox}
\end{table*}

\begin{table*}[h]
\centering
\caption{Design space of the power-to-methanol system}
\label{tab:design_space}
\begin{adjustbox}{max width=\textwidth}
\begin{tabular}{lccc}
\toprule
\textbf{Variable} & \textbf{Symbol} & \textbf{Unit} & \textbf{Range (energy equivalent)} \\
\midrule
Methanol production rate & $\dot{m}_{\text{MeOH}}$ & kg~h$^{-1}$ & [432.49, 2162.47] ([5,25MW]) \\
PEM electrolyzer capacity ratio & $\alpha_\text{PEM}$ & - & [0, 1]\\
Battery energy storage capacity & $C_\text{BESS}$ & MWh & [5, 100] \\
Compressed hydrogen storage tank capacity & $C_\text{CHT}$ & kg & [89.77, 1795.33] ([5,100 MWh]) \\
\bottomrule
\end{tabular}
\end{adjustbox}
\vspace{1mm}
\parbox{0.9\textwidth}{\raggedright\footnotesize
* Energy equivalents are calculated based on the specific power consumption for methanol production (11.56~kJ~kg$^{-1}$) and hydrogen (55.70~kJ~kg$^{-1}$) derived from \cref{tab:technicalparameter}}
\end{table*}

\begin{table*}[!t]
\centering
\caption{Hyperparameters of the proposed and benchmark models}
\label{tab:hyper-all}
\resizebox{\textwidth}{!}{
\begin{tabular}{lccccc}
\toprule
\textbf{Hyperparameters} 
& \textbf{BC} 
& \textbf{DRL(PPO)} 
& \textbf{BC+DRL(PPO)} 
& \textbf{DT} 
& \textbf{ST \& PT} \\
\midrule
\text{Sequence length $K$} 
& - & - & - & 24 & 24 \\
\text{Number of attention blocks} 
& - & - & - & 4 & 4 \\
\text{Embedding dimension} 
& - & - & - & 256 & 256 \\
\text{Hidden layer} 
& (256,256,256,128) & (256,256,256,128) & (256,256,256,128) & 256 & 256 \\
\text{Activation function} 
& \text{ReLU} & \text{ReLU} & \text{ReLU} & \text{ReLU} & \text{ReLU} \\
\text{Epochs} 
& 100 & 100 & 100 & 100 & 100\\
\text{Training batch size} 
& 128 & 128 & 128 & 64 & 64 \\
\text{Learning rate} 
& $1\!\times\!10^{-5}$ 
& $1\!\times\!10^{-5}$ 
& $1\!\times\!10^{-5}$ 
& $1\!\times\!10^{-4}$ 
& $1\!\times\!10^{-4}$ \\
\text{Weight decay} 
& $1\!\times\!10^{-4}$ 
& $1\!\times\!10^{-4}$
& $1\!\times\!10^{-4}$ 
& $1\!\times\!10^{-4}$ 
& $1\!\times\!10^{-4}$ \\
\textsc{Warmup steps} 
& - & - & - & 10,000 & 10,000 \\
\text{Optimizer} 
& Adam & Adam & Adam & AdamW & AdamW \cite{loshchilov2017decoupled} \\
\text{Hyperparaemter tunning} 
& - & PBT & PBT \cite{jaderberg2017population} & - & - \\
\bottomrule
\end{tabular}}
\vspace{1mm}
\parbox{0.9\textwidth}{\raggedright\footnotesize
Detailed model architecture is available at \url{https://github.com/PSEKJD/mascor/tree/main/mascor/models/agent}, and training implementations are available at \url{https://github.com/PSEKJD/mascor/tree/main/mascor/train}.}
\end{table*}

\begin{table*}[h]
\centering
\caption{CRPS of goal prediction (RTG and CTG) for ST and PT critic model under four design categories in offline and online planning}
\label{tab:crps}
\resizebox{\textwidth}{!}{
\begin{tabular}{l l cc cc}
\toprule
 &  
 & \multicolumn{2}{c}{\textbf{ST}} 
 & \multicolumn{2}{c}{\textbf{PT (proposed)}} \\
\cmidrule(r){3-4} \cmidrule(r){5-6}
 & \textbf{Design category} 
 & \textbf{RTG} & \textbf{CTG} & \textbf{RTG} & \textbf{CTG} \\
\midrule
\multirow{4}{*}{\textbf{Offline operation}}
 & Base        & 0.35$\pm$0.01 & 0.19$\pm$0.03 & \textbf{0.29$\pm$0.01} & \textbf{0.10$\pm$0.02} \\
 & Buffered     & 0.44$\pm$0.02 & 0.04$\pm$0.01 & \textbf{0.38$\pm$0.04} & \textbf{0.03$\pm$0.01} \\
 & Responsive   & 0.35$\pm$0.01 & 0.11$\pm$0.02 & \textbf{0.21$\pm$0.01} & \textbf{0.05$\pm$0.01} \\
 & Max-product  & 0.41$\pm$0.03 & 0.34$\pm$0.07 & \textbf{0.31$\pm$0.01} & \textbf{0.23$\pm$0.05} \\
\midrule
\multirow{4}{*}{\textbf{Online operation}}
 & Base        & 0.36$\pm$0.20 & 0.11$\pm$0.07 & \textbf{0.30$\pm$0.17} & \textbf{0.08$\pm$0.04} \\
 & Buffered     & 0.45$\pm$0.19 & 0.04$\pm$0.02 & \textbf{0.41$\pm$0.21} & \textbf{0.03$\pm$0.02} \\
 & Responsive   & 0.40$\pm$0.25 & 0.12$\pm$0.09 & \textbf{0.32$\pm$0.19} & \textbf{0.09$\pm$0.05} \\
 & Max-product  & 0.37$\pm$0.24 & 0.18$\pm$0.12 & \textbf{0.34$\pm$0.19} & \textbf{0.14$\pm$0.06} \\
\bottomrule
\end{tabular}}
\vspace{1mm}
\parbox{0.9\textwidth}{\raggedright\footnotesize
* Lower CRPS values indicate higher goal-prediction accuracy of the critic models. 
\\Bold scores denote design categories where the proposed PT achieves better stochastic goal-forecasting performance across diverse scenario.}
\end{table*}

\clearpage
\refstepcounter{section}
\section*{Algorithms}\label{sec:algorithm}
\begin{algorithm}
\caption{Actor–Critic offline operation}
\label{algo:offline}
\begin{algorithmic}[1]
\Require Actor $\pi_\theta$, Critic $Q_\theta'$, system model $\mathrm{ENV}$
\Require Design token $D$, Renewable trend token $E$, 
Operation horizon $T = 576$, initial state $s_{\text{init}}$, number of action samples $N$
\Require Trajectory buffer 
$\tau = \{(s_k, a_k, c_k, r_k, CTG_k, RTG_k)\}_{k=1}^{K}$, 
buffer size $K = 24$
\State Initialize $\tau$: set $s_1 = s_{\text{init}}$, and 
$s_k, a_k, c_k, r_k, CTG_k, RTG_k \leftarrow 0$ for $k>1$
\For{$t = 1$ \textbf{to} $T$}
    \State Sample $N$ candidate actions 
    $\{\hat{a}_t^i\}_{i=1}^N \sim \pi_\theta(\cdot \mid \{D \cup \tau\})$
    \State Collect candidate tuple 
    $(s_{t+1}^i, r_t^i, c_t^i) = \mathrm{ENV}(s_t, \hat{a}_t^i)$
    \State Sample long-term goal 
    $(\hat{RTG}_t^i, \hat{CTG}_t^i) = 
    Q_{\theta'}(\cdot \mid 
    \{\{D, E\} \cup \tau \cup 
    \{s_t, \hat{a}_t^i, c_t^i, r_t^i\}\})$
    \If{any feasible $i$ with $\hat{CTG}_t^{i} \le 0$ exists}
        \State $a_t^* = 
        \arg\max_{i=1,\dots,N} \hat{RTG}_t^{i} 
        \text{ s.t. } \hat{CTG}_t^{i} \le 0$
    \Else
        \State $a_t^* = 
        \arg\max_{i=1,\dots,N} \hat{RTG}_t^{i}$
    \EndIf
    \State Apply action $a_t^*$ and compute next state, reward, and carbon emission:
    \Statex \hspace{\algorithmicindent}
    $(s_{t+1}^*, r_t^*, c_t^*) = \mathrm{ENV}(s_t, a_t^*)$
    \State Update $RTG_{t+1} \leftarrow RTG_t^* - r_t^*$, \;
    $CTG_{t+1} \leftarrow CTG_t^* - c_t^*$
    \State Update trajectory 
    $\tau \leftarrow \tau \cup 
    \{(a_t^*, c_t^*, r_t^*, CTG_{t+1}, RTG_{t+1})\}$
\EndFor
\Statex \footnotesize Detailed implementation is available at \url{https://github.com/PSEKJD/mascor/blob/main/mascor/optimization/uq_problem.py}.
\end{algorithmic}
\end{algorithm}

\begin{algorithm}
\caption{Actor--Critic online operation}
\label{algo:online}
\begin{algorithmic}[1]
\Require Actor $\pi_\theta$, Critic $Q_\theta'$, system model $\mathrm{ENV}$
\Require Scenario generator $G$, discriminator $D$
\Require Scenario forecasting frequency $f_{\text{update}}$, projection operator 
$\mathbb{P}_{\text{hist}}\!\bigl(f(\mathbf{w})\bigr) = [w_1, \ldots, w_{h}]^{\mathsf{T}}$
with observation length $h$, optimization iteration number $N_z$
\Require Design token $D$, Operation horizon $T = 576$, 
initial state $s_{\text{init}}$, number of action samples $N$, 
number of scenario samples $M$
\Require Trajectory buffer 
$\tau = \{(s_k, a_k, c_k, r_k, CTG_k, RTG_k)\}_{k=1}^{K}$, 
buffer size $K = 24$

\State Initialize $\tau$: set $s_1 = s_{\text{init}}$, and 
$s_k, a_k, c_k, r_k, CTG_k, RTG_k \leftarrow 0$ for $k>1$
\State Initialize noise batch $\{z_j\}_{j=1}^{M} \sim \mathcal{N}(0, \Sigma)$
\For{$t = 1$ \textbf{to} $T$}
    \State Sample $N$ candidate actions $\{\hat{a}_t^i\}_{i=1}^N 
    \sim \pi_\theta(\cdot \mid \{D \cup \tau\})$
    \State Collect candidate tuple 
    $(s_{t+1}^i, r_t^i, c_t^i) = \mathrm{ENV}(s_t, \hat{a}_t^i)$
    \If{$t \bmod f_{\text{update}} = 0$}
        \State Optimize batch noise to align with 
        $w_{\text{obs}} = [w_{1}\cdots w_t]^\top$
        \For{$n_z = 1$ \textbf{to} $N_z$}
            \State Compute noise batch loss:
            \[
            \mathcal{L}_z = 
            \frac{1}{M}\sum_{j=1}^{M}
            \Biggl[
                \frac{1}{t}\bigl\|w_{\text{obs}} 
                - \mathbb{P}_{\text{hist}}\!\bigl(G(z_j)\bigr)\bigr\|_2^2
                - D\bigl(G(z_j)\bigr)
            \Biggr]
            \]
            \State Update noise batch using gradient descent:
            \[
            z_j \leftarrow z_j - \eta \, \nabla_{z_j}\mathcal{L}_z
            \]
            \If{$\mathcal{L}_z < 10^{-3}$}
                \State \textbf{break}
            \EndIf
        \EndFor
    \EndIf
    \State Generate predicted renewable scenarios 
    $w_{\text{pred}}^j = G(z_j)$
    \State Compute renewable trend token batch 
    $\{E_j\}_{j=1}^{M}$
    \State Propagate uncertainty of $\{E_j\}_{j=1}^{M}$:
    \[
    \hat{RTG}_t^i,\, \hat{CTG}_t^i =
    Q_\theta'(\cdot \mid 
    \{D \cup \{E_j\}_{j=1}^{M} \cup \tau 
    \cup \{s_t, \hat{a}_t^i, c_t^i, r_t^i\}\})
    \]
    \State Perform $a_t^*$ screening, $RTG_{t+1}$, $CTG_{t+1}$, 
    \Statex \hspace{\algorithmicindent}and $\tau$ update as defined in \cref{algo:offline}
\EndFor
\Statex \footnotesize Detailed implementation is available at \url{https://github.com/PSEKJD/mascor/blob/main/mascor/solvers/PT_solver.py}.
\end{algorithmic}
\end{algorithm}

\clearpage
\bibliographystyle{bst/sn-nature}
\bibliography{reference}

\end{document}